\documentclass[11pt]{article}

\usepackage[preprint]{acl}

\usepackage{times}
\usepackage{latexsym}

\usepackage[T1]{fontenc}
\usepackage[utf8]{inputenc}

\usepackage{microtype}

\usepackage{inconsolata}

\usepackage{graphicx}

\usepackage{booktabs}  % 必须，画三线表
\usepackage{multirow}  % 必须，合并行
\usepackage[table]{xcolor}
\usepackage{algorithm}
\usepackage{algpseudocode}
\usepackage{amsmath}
\usepackage{amssymb}
\usepackage{subcaption}

\usepackage{listings}
\usepackage{xcolor}

\lstdefinelanguage{Python}{
  keywords={False,None,True,and,as,assert,break,class,continue,def,del,elif,else,except,finally,for,from,global,if,import,in,is,lambda,nonlocal,not,or,pass,raise,return,try,while,with,yield},
  sensitive=true,
  morecomment=[l]\#,
  morestring=[b]',
  morestring=[b]"
}

\definecolor{forestgreen}{RGB}{34,139,34}
\usepackage{fvextra}  % 比 fancyvrb 更方便
\usepackage{booktabs}
\usepackage{pifont}
\usepackage{url}
\newcommand{\cmark}{\ding{51}}
\newcommand{\xmark}{\ding{55}}
\DefineVerbatimEnvironment{myverb}{Verbatim}{
  breaklines=true,     % 允许自动换行
  breakanywhere=true,  % 单词中间也可以断行（防止超宽）
  fontsize=\small      % 字体大小（可按需改）
}
\title{CreativeBench: Benchmarking and Enhancing Machine Creativity via Self-Evolving Challenges}
\author{
    Zi-Han Wang\textsuperscript{1,2,6}\thanks{Equal contribution. Work done during an internship at Southern University of Science and Technology.} 
    Lam Nguyen\textsuperscript{2}\footnotemark[1]
    Zhengyang Zhao\textsuperscript{3} \\
    \textbf{Mengyue Yang}\textsuperscript{4} 
    \textbf{Chengwei Qin}\textsuperscript{5} 
    \textbf{Yujiu Yang}\textsuperscript{2} 
    \textbf{Linyi Yang}\textsuperscript{1}\thanks{Corresponding author  <yangly6@sustech.edu.cn>.}
    \\
    \textsuperscript{1}Southern University of Science and Technology \\
    \textsuperscript{2}Tsinghua University 
    \textsuperscript{3}Peking University 
    \textsuperscript{4}University of Bristol \\
    \textsuperscript{5}The Hong Kong University of Science and Technology (Guangzhou)
    \textsuperscript{6}Xi’an Jiaotong University \\
    \texttt{zihanwang25@stu.xjtu.edu.cn, yangly6@sustech.edu.cn}
}
\begin{document}
\maketitle
\begin{abstract} 
The saturation of high-quality pre-training data has shifted research focus toward evolutionary systems capable of continuously generating novel artifacts, leading to the success of AlphaEvolve. However, the progress of such systems is hindered by the lack of rigorous, quantitative evaluation. To tackle this challenge, we introduce CreativeBench, a benchmark for evaluating machine creativity in code generation, grounded in a classical cognitive framework. Comprising two subsets -- CreativeBench-Combo and CreativeBench-Explore -- the benchmark targets combinatorial and exploratory creativity through an automated pipeline utilizing reverse engineering and self-play. By leveraging executable code, CreativeBench objectively distinguishes creativity from hallucination via a unified metric defined as the product of quality and novelty. Our analysis of state-of-the-art models reveals distinct behaviors: (1) scaling significantly improves combinatorial creativity but yields diminishing returns for exploration; (2) larger models exhibit ``convergence-by-scaling,'' becoming more correct but less divergent; and (3) reasoning capabilities primarily benefit constrained exploration rather than combination. Finally, we propose EvoRePE, a plug-and-play inference-time steering strategy that internalizes evolutionary search patterns to consistently enhance machine creativity. We release our data and code at: \href{https://zethwang.github.io/creativebench.github.io/}{CreativeBench Homepage}.

\end{abstract}
\section{Introduction}

The success of Large Language Models has been driven by scaling up Internet-scale datasets~\citep{NEURIPS2020_1457c0d6,han2021pretrainedmodelspastpresent,xu2025probingscientificgeneralintelligence}. However, this approach now faces a bottleneck: the saturation of high-quality web data limits further scaling of model intelligence. This limitation has renewed interest in \emph{evolutionary systems}~\citep{borg2022evolvedopenendednessculturalevolution, faldor2024artificialopenendedevolutionlenia}, which are intended to continually produce artifacts that remain both \emph{novel} and \emph{learnable}~\citep{hughes2024openendednessessentialartificialsuperhuman}.
Evolving systems
% \footnote{evolving systems focus on the continuous generation of novel artifacts, while open-ended tasks derive openness from inherently subjective or ill-defined objectives.} 
typically instantiate this paradigm by pairing foundation models (as rich conceptual priors) with evolutionary algorithms (as mechanisms for exploration)~\citep{wang2023voyageropenendedembodiedagent, romera-paredes2024mathematical,novikov2025alphaevolvecodingagentscientific}.

% \begin{figure}
%     \centering
%     \includegraphics[width=1\linewidth]{58f116a5ad10396c855da75c7968b8d7.png}
%     \caption{Enter Caption}
%     \label{fig:placeholder}
% \end{figure}

\captionsetup[subfigure]{font=footnotesize}
\begin{figure}[t]
    \centering
    \begin{subfigure}{0.49\columnwidth}
        \centering
        \includegraphics[width=\linewidth]{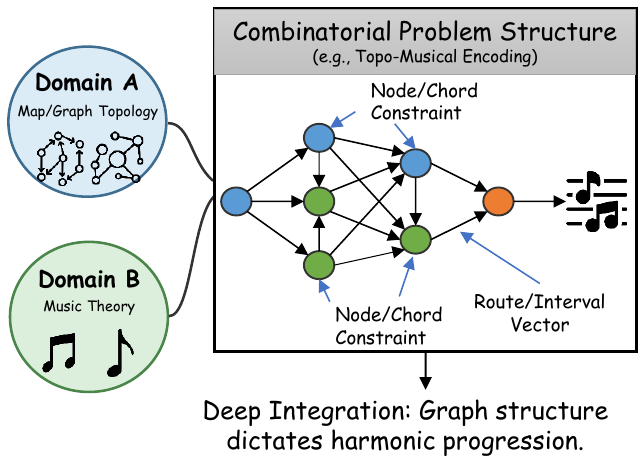}
        \caption{\small Combinatorial Creativity}
        \label{case_figure1_combinatorial}
    \end{subfigure}
    \hfill
    \begin{subfigure}{0.49\columnwidth}
        \centering
        \includegraphics[width=\linewidth]{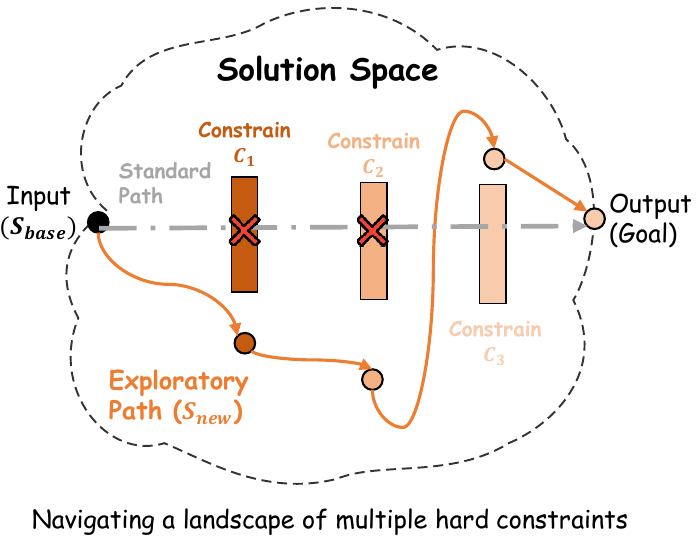}
        \caption{Exploratory Creativity}
        \label{case_figure2_exploratorial}
    \end{subfigure}
    \caption{The demonstration of two types of machine creativity considered in CreativeBench.}
    % \caption{Overall caption describing both subfigures.}
    \label{case_figures}
\end{figure}

While promising, the progress of these systems is currently hindered by the lack of rigorous measurement~\citep{lehman2011abandoning,lange2023neuroevobench}. Existing works prioritize functional correctness, overlooking the direct evaluation of creativity. Even when creativity is considered, evaluations often (1) struggle to distinguish creativity from hallucination~\citep{sui-etal-2024-confabulation,jiang2024surveylargelanguagemodel} objectively, (2) lack sufficient task complexity to elicit truly creative behaviors rather than rote memorization~\citep{delorenzo2024creativevalevaluatingcreativityllmbased,lu-etal-2025-benchmarking}, and (3) lack grounded, automatable quantitative metrics for creativity in evolving systems.

To bridge these gaps, we adopt the cognitive creativity framework proposed by Boden~\citep{Boden2004}. This framework categorizes creativity into distinct types, including \emph{combinatorial creativity} -- combining familiar concepts in unfamiliar ways, -- and \emph{exploratory creativity} -- navigating a structured conceptual space to discover new possibilities (see Figure~\ref{case_figures}). Accordingly, we introduce \textbf{CreativeBench}, a benchmark for code generation systems comprising two subsets, CreativeBench-Explore and CreativeBench-Combo, which focus on exploratory creativity and combinatorial creativity, respectively, based on the assumption that these two types of machine creativity capture the two core capabilities required for evolutionary systems: constraint-driven search and recombining concepts.

To this end, we present a timely dataset from several perspectives.
First, unlike creative writing~\citep{paech2024eqbenchemotionalintelligencebenchmark,wu2025writingbenchcomprehensivebenchmarkgenerative}, code utilizes objective execution to strictly distinguish creativity from hallucination. Second, to ensure task complexity while reducing confounds from data leakage and rote memorization, we build an automated pipeline, constructing a high-difficulty benchmark to guarantee that derived tasks genuinely reflect creative behaviors rather than memorization, with Pass@1 remaining below 60\% even for Gemini-3-Pro~\citep{gemini3pro}. Third, to ensure evaluation metrics are grounded, we define a quantitative \emph{creativity score} as the product of \emph{Quality} and \emph{Novelty}~\citep{williams1980cap_manual}. Quality is verified via sandboxed execution and LLM-as-a-judge, while novelty is measured by the logic distance between candidate programs and appropriate baselines~\citep{runco2012standard}. Finally, we invite human experts to verify both data quality and metric reliability, achieving 89.1\% instance validity and strong agreement between automated and human creativity rankings (Spearman's $\rho=0.78$).

Building on CreativeBench, we analyze state-of-the-art foundation models along with evolutionary algorithms and highlight three key \textbf{insights}:
(i) \emph{Scaling Favors Combination over Exploration}: scaling substantially improves combinatorial creativity, yet yields limited  gains for exploratory creativity;
(ii) \emph{Convergence-by-Scaling}: larger models are more correct but less divergent;
(iii) \emph{Reasoning Helps Exploration, Not Combinatorial Creativity}: reasoning primarily benefit constraint-driven exploration rather than cross-domain combination.

Beyond these findings, we propose \textbf{EvoRePE} (Evolutionary Representation Engineering), a plug-and-play inference-time steering method that extracts a creativity vector from evolutionary trajectories. EvoRePE yields creativity gains that are orthogonal to the underlying evolutionary strategy, suggesting that part of evolutionary optimization can be internalized as latent-space steering toward a steered evolution paradigm.

To our knowledge, we are the first to contribute a machine creativity benchmark based on Boden's cognitive creativity framework \citep{Boden2004}. Our contributions are three-fold as follows:

\begin{itemize}
    \item We construct a benchmark measuring machine creativity by considering both \emph{exploratory} and \emph{combinatorial} creativity.
    \item We uncover insights into how model scaling and reasoning capabilities interact with creativity in evolutionary systems by proposing a novel evaluation metric.
    \item We propose EvoRePE, a plug-and-play method that effectively enhances model creativity by steering latent representations.
\end{itemize}

\begin{figure*}
    \centering
    \includegraphics[width=1\linewidth]{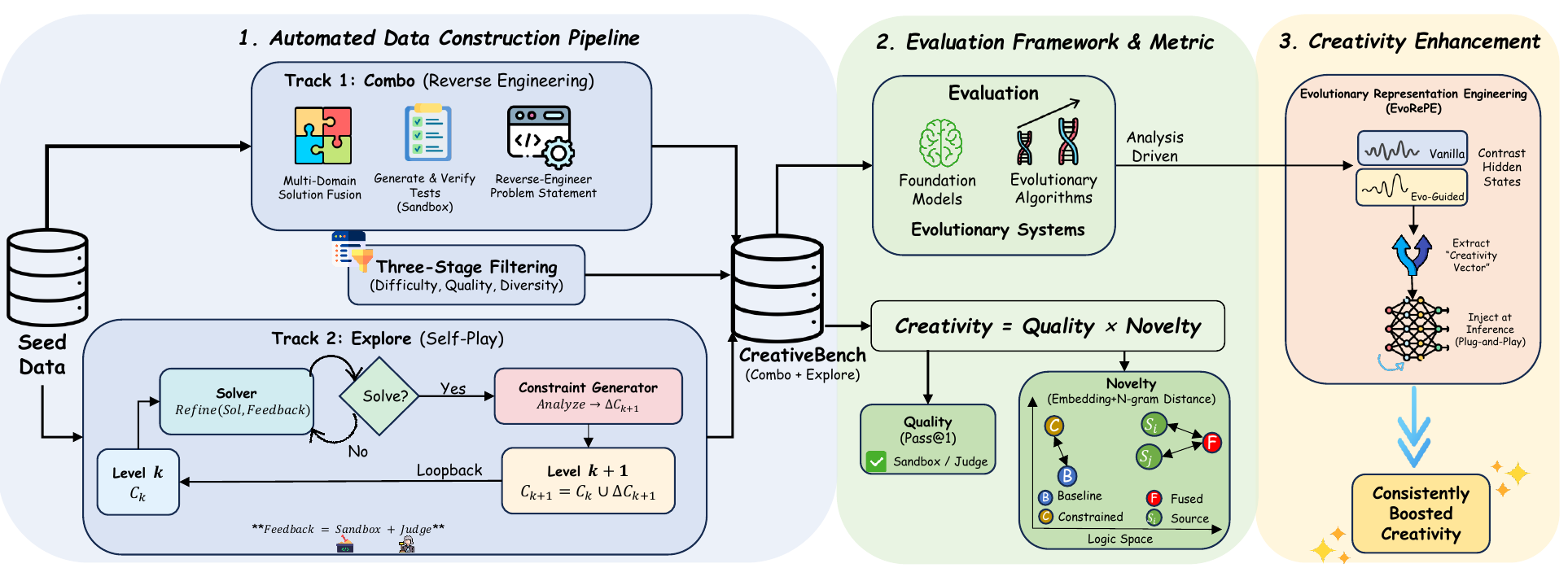}
    \caption{\label{fig:overview}
    \textbf{Overview of our framework.} 
    \textbf{(Left)} We introduce \textbf{CreativeBench}, built via an automated reverse engineering and self-play pipeline. 
    \textbf{(Middle)} We evaluate evolutionary systems using a unified \textbf{Creativity Score}, defined as the  \emph{Quality} (Pass@1) and \emph{Novelty} (embedding + n-gram distance). 
    \textbf{(Right)} Based on our analysis, we propose the \textbf{EvoRePE} strategy to steer models toward more creative solutions at inference time.}
    \label{fig:framework}
\end{figure*}

\begin{table}[t]
\centering
\setlength{\tabcolsep}{1.5pt}
\resizebox{\columnwidth}{!}{%
\begin{tabular}{lccccccccc}
\toprule
\textbf{Benchmark} & \textbf{Metric} & \textbf{\#Prob.} & \textbf{Creativity} & \textbf{Explore} & \textbf{Combo} & \textbf{Auto.} & \textbf{Difficulty} & \textbf{Domain} & \textbf{Len} \\
\midrule
HumanEval~\citep{chen2021evaluatinglargelanguagemodels} & Pass@$k$ & 164 & \xmark & \xmark & \xmark & \xmark & $\star$ & 5 & 134 \\
MBPP~\citep{austin2021programsynthesislargelanguage} & Pass@$k$ & 974 (500 test) & \xmark & \xmark & \xmark & \xmark & $\star$ & 6 & 50 \\
LiveCodeBench~\citep{jain2024livecodebenchholisticcontaminationfree} & Pass@$k$ & 400--880 & \xmark & \xmark & \xmark & \xmark & $\star\star\star\star$ & 4 & 470 \\
EvoCodeBench~\citep{NEURIPS2024_6a059625} & Pass@$k$/Recall@$k$ & 275 & \xmark & \xmark & \xmark & \cmark & $\star\star\star$ & 10 & 132 \\

CreativeEval~\citep{delorenzo2024creativevalevaluatingcreativityllmbased} & FFOE & 120 & \cmark & \xmark & \xmark & \xmark & - & 1 & - \\

NeoCoder~\citep{lu-etal-2025-benchmarking} & Divergent/Convergent & 199 & \cmark & \xmark & \xmark & \xmark & $\star\star$ & 1 & 494 \\

\midrule
\textbf{CreativeBench} & \textbf{Quality$\times$Novelty} & \textbf{1,859} & \cmark & \cmark & \cmark & \cmark & $\star\star\star\star\star$ & \textbf{14} & \textbf{593} \\
\bottomrule
\end{tabular}%
}
\caption{Comparison of existing code generation benchmarks and ours.
\textbf{\#Prob.}: Number of problems in the commonly used setting (LiveCodeBench varies across snapshots).
\textbf{Explore}: Evaluates exploratory creativity via negative constraints.
\textbf{Combo}: Requires domain fusion (combinatorial).
\textbf{Auto.}: Fully automated data construction pipeline (human-free).
\textbf{Len}: Average token length of problem descriptions.
\textbf{FFOE}: Fluency, Flexibility, Originality, and Elaboration.}
\end{table}

\label{tab:comparisons}

\section{Related Work}
\paragraph{Machine Creativity Evaluation.}

While human creativity has been extensively studied in psychological and cognitive science~\citep{Amabile1982social,mumford1991process,guilford1950creativity}, the evaluation of creativity remains underexplored. In the task of code generation, models are predominantly judged by functional correctness via Pass@k metrics~\citep{chen2021evaluatinglargelanguagemodels,jimenez2024swebench,NEURIPS2024_6a059625}, often ignoring the creative dimensions of problem-solving. Recent work evaluates LLM creativity through divergent and convergent thinking~\citep{bok2025reasoningobviousevaluatingdivergent,sen2025think,lu-etal-2025-benchmarking}. We compare our benchmark with prior methods in Table~\ref{tab:comparisons}.
In particular, we show that existing work evaluating LLM creativity typically has three major drawbacks: (1) reliance on subjective assessments that struggle to distinguish creativity from hallucination~\citep{jiang2024surveylargelanguagemodel,sui-etal-2024-confabulation,Wang2025,atmakuru2024cs4measuringcreativitylarge}; (2) insufficient task complexity to elicit truly creative solutions rather than memorization~\citep{delorenzo2024creativevalevaluatingcreativityllmbased,Zhao_2025,jiang2025artificial}; and (3) the lack of reliable, grounded quantitative measurements for evolutionary systems. 

\paragraph{Evolutionary Algorithm.}

Pursuing machine creativity aims to build systems that can continually generate artifacts that are both \emph{novel} and \emph{learnable} to an observer \citep{hughes2024openendednessessentialartificialsuperhuman}. To achieve this goal, recent research has moved beyond purely static training paradigms toward mechanism-driven approaches using self-evolution methods. FunSearch \citep{romera-paredes2024mathematical} combines LLMs with a programmatic evaluator to search for mathematical solutions in function space. AlphaEvolve \citep{novikov2025alphaevolvecodingagentscientific} further systematizes this by employing island-style genetic search to maintain population diversity during code evolution. Similarly, GEPA \citep{agrawal2025gepareflectivepromptevolution} treats prompts as evolvable genotypes, using multi-objective search to optimize instructions. Our work builds on these code-centric evolutionary strategies but focuses specifically on evaluating the \emph{creativity} of the generated solutions.
\paragraph{Representation Engineering.}

Representation engineering controls LLM behavior by monitoring and intervening on residual streams and internal activations \citep{zou2025representationengineeringtopdownapproach, turner2024steeringlanguagemodelsactivation}. A common approach is to extract a steering vector from the difference in activations between opposing pairs (e.g., honest vs.\ dishonest, neutral vs.\ biased) \citep{rimsky-etal-2024-steering}.
Existing research has predominantly applied representation engineering to alignment tasks, such as enhancing truthfulness~\citep{li2023inferencetime}, mitigating bias~\citep{siddique2025shiftingperspectivessteeringvectors, rimsky-etal-2024-steering}, or controlling emotional style~\citep{konen-etal-2024-style, pai2025billysteeringlargelanguage}. However, it remains unclear whether vector perturbations can effectively enhance the \emph{creativity} of LLMs and be combined with evolutionary algorithms. We show that a latent ``creativity direction'' can be extracted via evolutionary prompting, and the resulting vector serves as a plug-and-play signal to steer models toward creative solutions.
\begin{table}[t]
\centering
% 将表格宽度强制设置为当前栏宽 (\columnwidth)
\resizebox{\columnwidth}{!}{%
    \begin{tabular}{lccccc} % 注意：你原本写了7个c，但实际只有6列数据，这里改成了6个
    \toprule
    Dataset & \#Problems & \#Test Cases & Prob Len & Solu Len & Domain \\
    \midrule
    CreativeBench-Combo & 1308 & 16404 & 593.0 & 776.0 & 14 \\
    CreativeBench-Explore   & 551  & 8452  & 268.0 & 171.0 & 14 \\
    \bottomrule
    \end{tabular}%
}
\caption{Statistics of CreativeBench.}
\label{tab:stats}
\end{table}

\section{CreativeBench}

\subsection{Overview} 
%CreativeBench is a large-scale, high-difficulty code generation benchmark with two complementary subsets: CreativeBench-Combo and CreativeBench-Explore. Combinatorial tasks require solving a new problem by integrating logic from multiple source domains, while exploratory tasks require producing a functionally correct solution that additionally satisfies strict negative constraints. 

We build CreativeBench using GPT-4.1~\citep{gpt4.1}, taking the full Python subset of AutoCodeBench as seed tasks (196 problems)~\citep{chou2025autocodebenchlargelanguagemodels}. As shown in Table~\ref{tab:stats}, CreativeBench spans 14 domains, providing broad coverage of programming scenarios. To maintain task complexity while mitigating data leakage and rote memorization, we build CreativeBench with a scalable \emph{reverse engineering} and \emph{self-play} pipeline (Figure~\ref{fig:framework}).

\subsection{CreativeBench-Combo Dataset}

For CreativeBench-Combo, we adopt a reverse-engineering strategy that derives problem descriptions from pre-validated composite code to generate the synthesis of high-difficulty combinatorial tasks that are inherently solvable.

\paragraph{Solution Fusion.} We first prompt the model to combine code components from different domains (e.g., merging data processing with graph algorithms) into a single, unified solution. These candidates are executed in a sandbox to strictly verify their correctness. This ``code-first'' approach guarantees that every generated task comes with a verified reference solution.

\paragraph{Test Function Generation.} To enable objective evaluation, we generate test cases directly from the verified solution. We instruct the model to create valid inputs, which are then run in the sandbox to obtain the ground-truth outputs. These input-output pairs are formatted into standard assertion statements to construct the final test function.

\paragraph{Problem Synthesis.} Finally, we reconstruct the problem description. We ask the model to interpret the semantic intent of the code: \emph{Given this solution and its test cases, what problem does it solve?} To ensure the description is high-quality, we provide a set of strict guidelines that the model must follow to synthesize a clear and coherent problem statement.

\subsection{CreativeBench-Explore Dataset}

We employ a self-play construction method based on the asymmetry that \emph{creating a constraint is easier than solving a problem under that constraint}. We structure the process as a dynamic interaction between a \emph{Constraint Generator} and a \emph{Solver}, where the difficulty increases progressively.

\paragraph{Dynamic Constraint Stacking.}
We generate constraints in iterative levels. Starting from the unconstrained problem (Level~0), the process advances to Level~$k{+}1$ only when the instance at Level~$k$ is successfully solved. At each step, the Generator analyzes the Solver's solution from the previous level and introduces a new \emph{negative constraint} to invalidate its specific algorithmic choices (e.g., forbidding a particular operator or control-flow pattern). The new constraint is stacked onto the existing set $\mathcal{C}_k$, yielding
$\mathcal{C}_{k+1} = \mathcal{C}_k \cup \{\Delta \mathcal{C}_{k+1}\}$,
so that higher levels strictly subsume all prior restrictions. This mechanism continuously pushes the Solver toward structurally distinct algorithms until it reaches its capability limit.

\paragraph{Refinement and Termination.}
To determine whether a level is valid (i.e., solvable), the Solver attempts to satisfy the accumulated constraints $\mathcal{C}_k$ via a reference-guided refinement strategy. As shown in Eq.~\ref{eq:refine}, the Solver iteratively modifies a base solution $S_{\mathrm{base}}$ using feedback from sandbox and judger:
\begin{equation}
\label{eq:refine}
\small
S_{\mathrm{new}}
\leftarrow
\mathrm{Refine}\!\left(
S_{\mathrm{current}},\, \mathrm{Feedback}
\;\middle|\;
S_{\mathrm{base}},\, \mathcal{C}_k
\right).
\end{equation}
For each level, the Solver is allowed up to a fixed number of refinement attempts (e.g., 3). If it produces a valid solution that passes both sandbox execution (correctness) and an LLM judge (constraint adherence), the newly added constraint is deemed effective and the process advances to the next level. Otherwise, if the Solver fails within the allotted attempts, we treat the current constraint set as too strict (or unsatisfiable for the model) and terminate the generation loop for this problem.

\subsection{Data Filtering}
We apply a three-stage filter to maintain benchmark quality. More details about prompts and evaluation criteria are provided in Appendix~\ref{app:algorithms}.
\paragraph{Difficulty Check.} Programming problems that are too simple are not meaningful for evaluating the
code generation capabilities of current LLMs. For each problem, we sample five answers with GPT-4 and validate them. Problems solved correctly in all attempts are removed. 
\paragraph{Quality Audit.}
GPT-4o audits each sample based on the same specification rules used in our dataset construction pipeline, performing an additional verification of problem quality. This includes checking the clarity of the problem statement, the correctness and completeness of the test function, and the consistency between the specification, reference solution, and executable tests.
\paragraph{Diversity Check.} To prevent redundancy and ensure broad conceptual coverage, we employ semantic de-duplication. We compute vector embeddings for all problem descriptions using \texttt{text-embedding-3-small}~\citep{text-embedding-3-small} and calculate pairwise cosine similarities. Pairs exceeding a strict similarity threshold (0.85) are flagged.

\subsection{Manual Verification}

In our automated pipeline, we use well-designed problem specifications and an LLM-as-Judge module to maintain quality control. However, since LLMs cannot guarantee 100\% accuracy, the overall quality of CreativeBench remains uncertain. To measure its reliability, we invited three expert annotators to perform a manual review.

We built a visualization interface that displays each record, including the problem statement, the test function, and the reference solution. Annotators checked whether the test function was correct and aligned with the problem description, and then assigned a binary label (\emph{yes/no}) indicating whether the instance was valid.
We randomly selected 100 samples from CreativeBench-Explore and 200 from CreativeBench-Combo. The review showed a data validity rate of 89.1\%, indicating that our automated construction process is reliable. In comparison, Gemini-2.5-Pro reached pass rates of 53.8\% and 48.2\% on these tasks, suggesting substantial room for improvement.

\begin{figure*}
    \centering
    \includegraphics[width=1\linewidth]{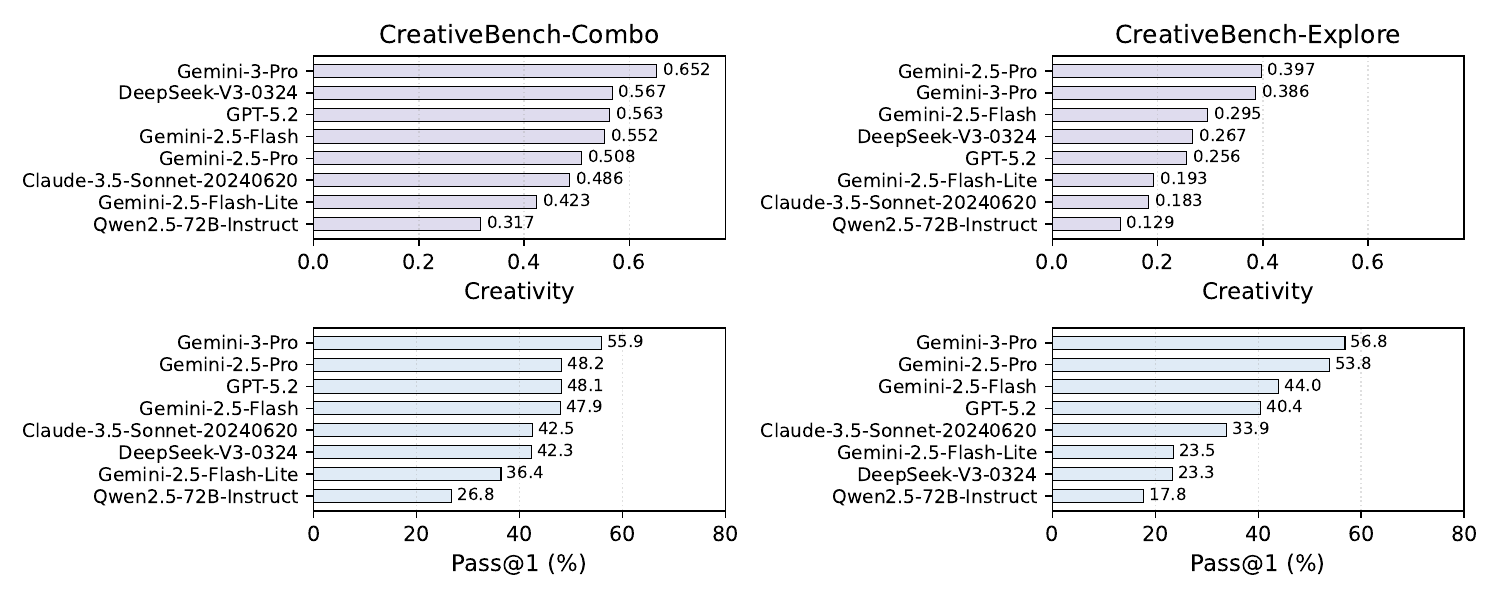}
    \caption{Performance of foundation models on CreativeBench.The left and right columns correspond to the Combinatorial (CreativeBench-Combo) and Exploratory (CreativeBench-Explore) subsets, respectively.}
    \label{fig:foundation_model}
\end{figure*}

\section{Experimental Setup}
\subsection{CreativeBench}
\paragraph{Models.}
We evaluate range of models, including Gemini-3-Pro~\citep{gemini3pro}, GPT-5.2~\citep{openai2025gpt52}, Gemini-2.5-Pro~\citep{gemini2.5}, Claude-3.5-Sonnet~\citep{claude3.5}, DeepSeek-V3~\citep{deepseek_v3}, Qwen3-4B-Instruct, Qwen3-8B-Instruct~\citep{yang2025qwen3}(with thinking mode on/off), Gemini-2.5-Flash-Lite~\citep{gemini2.5}, Qwen2.5-Instruct series (1.5B, 3B, 7B, 14B, 32B, 72B)~\citep{qwen2025qwen25technicalreport}. 
\paragraph{Baselines.}
We consider standard zero-shot prompting and two evolutionary optimization baselines, AlphaEvolve~\citep{novikov2025alphaevolvecodingagentscientific} and GEPA~\citep{agrawal2025gepareflectivepromptevolution}, using Gemini-2.5-Pro as the backend for iterative prompt mutation and feedback. AlphaEvolve employs an island-style genetic search framework that evolves candidate programs through iterative mutation and selection while maintaining population diversity. However, although AlphaEvolve is capable of operating directly at the program optimization level, the large scale of our dataset and the associated computational cost make such usage impractical. For efficiency, we therefore apply AlphaEvolve in a prompt optimization setting for evaluation.
GEPA treats prompts as evolvable genotypes and performs multi-objective evolutionary optimization with Pareto candidate selection, making it possible to retain complementary improvements across different behavioral dimensions when refining instructions. Details are provided in Appendix~\ref{sec:setup}.

% \paragraph{EvoRePE.}
% We compute a creativity vector by applying Principal Component Analysis (PCA)~\citep{shlens2014tutorialprincipalcomponentanalysis} to contrastive hidden states extracted from vanilla generations and AlphaEvolve-guided generations. During inference, this vector is added to the model’s hidden activations with a steering coefficient of $0.1$, guiding the model toward more creative solution patterns independent of the prompting strategy. In our implementation, the creativity vector is obtained from layer~26 of Qwen2.5-7B. 

\subsection{Evaluation Metrics}
\paragraph{Unified Creativity Score.}
%We evaluate the creativity
% \footnote{
% We focus on Psychological Creativity—novelty relative to the model's own prior knowledge—rather than Historical Creativity.
% See Appendix~\ref{app:theoretical_grounding} for a detailed discussion.}
%of models by measuring their ability to continually generate artifacts that preserve novelty and learnability. 

We define creativity as the expected product of quality and novelty across samples. The multiplicative formulation induces \emph{selectivity}: a solution receives a high creativity score only when it is both correct and meaningfully different from baseline solutions. Consequently, solutions that are correct but routine, or novel but incorrect, are assigned low creativity scores.
\begin{equation}
\small
\mathrm{Creativity}
=
\mathbb{E}_{i}\!\left[\mathrm{Quality}_i \times \mathrm{Novelty}_i\right].
\end{equation}

% \paragraph{Quality.}
% In practice, we adopt $\mathrm{Quality}_i = \mathrm{Pass@k}_i$ as the quality metric for code generation. Unless otherwise specified, we report $\mathrm{Pass@1}_i$, which can be viewed as a lower bound on the model’s ability to generate high-quality solutions. All generated solutions are executed and validated inside a sandbox~\citep{zhang2024autocodebench}.

\paragraph{Quality.} We measure solution quality by execution correctness and instantiate the metric as $\mathrm{Pass@1}$ which provides a lower-bound estimate of a model’s success probability under single-sample decoding. All generated solutions are executed and validated inside a sandbox~\citep{chou2025autocodebenchlargelanguagemodels}.

\paragraph{Novelty.}
We define novelty as the extent to which a solution differs from previously observed or baseline solutions, capturing the degree of originality or non-trivial departure from known patterns.
Furthermore, we quantify \textit{Novelty} by measuring the degree to which a generated solution deviates from a baseline solution within the solution space~\citep{Lehman2011,10.3389/frobt.2016.00040}. 
%Novelty in code is fundamentally rooted in structural, algorithmic, and logical differences, such as adopting different control flow paradigms, employing alternative algorithms, or altering the underlying logical execution pathways~\citep{novikov2025alphaevolvecodingagentscientific}, rather than superficial textual variations like variable renaming or code formatting changes~\citep{feser2022metricprogramsynthesis}. 
We adopt CodeXEmbed~\citep{liu2025codexembed}, a generalist code embedding model explicitly optimized to capture program structure and dependencies while remaining robust to syntactic variations. To further penalize near-copy behaviors with only lightweight textual edits, we complement embedding distance with a character-level n-gram distance. Concretely, for two solutions $u$ and $v$, let $\mathbf{e}_u$ and $\mathbf{e}_v$ denote embeddings, $\cos(\mathbf{e}_u,\mathbf{e}_v)$ the cosine similarity, and $G_4(\cdot)$ the set of distinct character 4-grams extracted from a solution. We instantiate the novelty as
\begin{equation}
\small
\mathcal{N}(u,v)
=
\underbrace{1-\cos(\mathbf{e}_u,\mathbf{e}_v)}_{\text{embedding}}
+
\underbrace{\left(1-\frac{|G_4(u)\cap G_4(v)|}{|G_4(u)\cup G_4(v)|}\right)}_{\text{4-gram}},
\end{equation}

\textbf{Exploratory Novelty} measures the deviation of a constrained solution $y_c$ from an unconstrained baseline $y_b$:
\begin{equation}
\small
\mathcal{N}_{\text{explore}} = \mathcal{N}(y_c, y_b).
\end{equation}

\textbf{Combinatorial Novelty} measures how much the combined-problem solution $y_c$ deviates from its $k$ source solutions $\{y_{s}^{(j)}\}_{j=1}^{k}$:
\begin{equation}
\small
\mathcal{N}_{\text{combo}} 
= 
\frac{1}{k}\sum_{j=1}^{k} \mathcal{N}\bigl(y_c, y_{s}^{(j)}\bigr).
\end{equation}
This encourages genuine integration rather than copying a single source with minor edits.
We provide additional robustness checks for \text{Novelty} under superficial edits and length in Appendix~\ref{app:novelty_robustness}. $\text{Quality}\in[0,1]$ and $\text{Novelty}\in[0,3]$, the Creativity score ($\text{Quality}\times\text{Novelty}$) is bounded in $[0,3]$.

\paragraph{Manual Validation.}
To verify the efficacy of our metrics, we conduct a manual review comprising two aspects. We compared our automated creativity rankings with expert rankings on a sampled subset. The results show a high consistency rate (Spearman's $\rho=\textbf{0.78}$), confirming that our metric reliably reflects perceived creativity. A detailed case study can be found in Appendix~\ref{app:case_study}.

\subsection{EvoRePE}

%Previous work points out that evolutionary algorithms significantly boost open-ended creativity but incur high computational costs due to iterative sampling~\citep{li2024evocodebenchevolvingcodegeneration}. 
We propose \textbf{EvoRePE} (Evolutionary Representation Engineering), a training-free strategy that distils the creative shifts found by evolutionary search into a compact steering vector.

\paragraph{Method.}
EvoRePE extracts a latent direction that captures the transition from a standard solution to an evolved solution.
Let $\mathcal{D}=\{(x_{\text{base}}^{(i)},x_{\text{evo}}^{(i)})\}_{i=1}^{N}$ be a dataset of $N$ prompt pairs, where $x_{\text{base}}^{(i)}$ is the initial standard prompt and $x_{\text{evo}}^{(i)}$ is the corresponding optimized prompt derived from an evolutionary algorithm (e.g., AlphaEvolve).
For a given layer $\ell$, let $\mathbf{h}_\ell(x)$ denote an aggregated activation vector for input $x$ (e.g., the last-token activation).
We compute per-pair shifts $\Delta \mathbf{h}_\ell^{(i)}=\mathbf{h}_\ell(x_{\text{evo}}^{(i)})-\mathbf{h}_\ell(x_{\text{base}}^{(i)})$ and collect them into a matrix $H_\ell$. We define the creativity vector as the Principal Component Analysis (PCA)~\citep{shlens2014tutorialprincipalcomponentanalysis}:
\[
\mathbf{v}_\ell = \mathrm{PCA}_1(H_\ell).
\]
During inference, we steer the residual stream by
\[
\tilde{\mathbf{h}}_\ell=\mathbf{h}_\ell+\alpha\,\mathbf{v}_\ell,
\]
where $\alpha$ controls the intervention strength.

\begin{table*}[!ht]
    \centering
    \renewcommand{\arraystretch}{1.2} 
    \setlength{\tabcolsep}{3pt}
    \resizebox{0.98\textwidth}{!}{%
        \begin{tabular}{cl r@{\hspace{2pt}}l r@{\hspace{2pt}}l r@{\hspace{2pt}}l | r@{\hspace{2pt}}l r@{\hspace{2pt}}l r@{\hspace{2pt}}l}
            \toprule
            \multirow{2}{*}{\textbf{Type}} &
            \multirow{2}{*}{\textbf{Method}} 
            & \multicolumn{6}{c|}{\textbf{CreativeBench-Combo}} 
            & \multicolumn{6}{c}{\textbf{CreativeBench-Explore}} \\
            \cmidrule(lr){3-8} \cmidrule(lr){9-14}
            && \multicolumn{2}{c}{\textbf{Pass@1 $\uparrow$}} & \multicolumn{2}{c}{\textbf{Novelty $\uparrow$}} & \multicolumn{2}{c}{\textbf{Creativity $\uparrow$}} 
            & \multicolumn{2}{c}{\textbf{Pass@1 $\uparrow$}} & \multicolumn{2}{c}{\textbf{Novelty $\uparrow$}} & \multicolumn{2}{c}{\textbf{Creativity $\uparrow$}} \\
            
            \midrule
            \rowcolor{gray!10}\multicolumn{14}{c}{\textbf{\textit{Qwen2.5-7B-Instruct}}}\\
            \midrule
            
            % --- Standard ---
            \multirow{2}{*}{Standard} 
                & Vanilla Prompt         
                & 9.21\% {\scriptsize$\pm$ 0.11\%} &       
                & 1.56 {\scriptsize$\pm$ 0.005} &      
                & 0.168 {\scriptsize$\pm$ 0.003} &       
                & 4.11\% {\scriptsize$\pm$ 0.18\%} &       
                & 0.473 {\scriptsize$\pm$ 0.006} &    
                & 0.0146 {\scriptsize$\pm$ 0.0005} & \\
                
                & \quad + \textbf{EvoRePE} {\footnotesize{(Ours)}} 
                & 9.71\% {\scriptsize$\pm$ 0.09\%} &       
                & 1.59 {\scriptsize$\pm$ 0.01} &       
                & 0.174 {\scriptsize$\pm$ 0.003} &       
                & 4.38\% {\scriptsize$\pm$ 0.15\%} &       
                & 0.469 {\scriptsize$\pm$ 0.006} &     
                & 0.0148 {\scriptsize$\pm$ 0.0006} & \\

            \cmidrule{1-14} 
            
            % --- Evolutionary ---
            \multirow{4}{*}{Evolutionary} 
                & AlphaEvolve            
                & 10.81\% {\scriptsize$\pm$ 0.08\%} &      
                & 1.53 {\scriptsize$\pm$ 0.01} &      
                & 0.175 {\scriptsize$\pm$ 0.003} &       
                & 5.22\% {\scriptsize$\pm$ 0.12\%} &       
                & 0.458 {\scriptsize$\pm$ 0.009} &    
                & 0.0163 {\scriptsize$\pm$ 0.0004} & \\
                
                & \quad + \textbf{EvoRePE}         
                & \textbf{11.48\%} {\scriptsize$\pm$ 0.11\%} &       
                & 1.57 {\scriptsize$\pm$ 0.01} &       
                & \textbf{0.193} {\scriptsize$\pm$ 0.003} &       
                & \textbf{5.75\%} {\scriptsize$\pm$ 0.18\%} &       
                & 0.457 {\scriptsize$\pm$ 0.009} &    
                & 0.0169 {\scriptsize$\pm$ 0.0006} & \\

                \cmidrule(lr){2-14} 
                
                & GEPA                   
                & 11.24\% {\scriptsize$\pm$ 0.09\%} &      
                & 1.54 {\scriptsize$\pm$ 0.01} &      
                & 0.176 {\scriptsize$\pm$ 0.003} &       
                & 4.65\% {\scriptsize$\pm$ 0.15\%} &       
                & 0.465 {\scriptsize$\pm$ 0.007} &    
                & 0.0162 {\scriptsize$\pm$ 0.0007} & \\
                
                & \quad + \textbf{EvoRePE}         
                & 11.47\% {\scriptsize$\pm$ 0.07\%} &       
                & 1.56 {\scriptsize$\pm$ 0.01} &       
                & 0.188 {\scriptsize$\pm$ 0.002} &       
                & 5.20\% {\scriptsize$\pm$ 0.10\%} &       
                & 0.470 {\scriptsize$\pm$ 0.008} &       
                & \textbf{0.0182} {\scriptsize$\pm$ 0.0006} & \\
                
            \midrule
            \rowcolor{gray!10}\multicolumn{14}{c}{\textbf{\textit{Gemini-2.5-Flash-Lite}}}\\
            \midrule
            Standard & Vanilla Prompt     
            & 36.41\% {\scriptsize$\pm$ 0.10\%} & 
            & 1.64 {\scriptsize$\pm$ 0.01} & 
            & 0.509 {\scriptsize$\pm$ 0.003} & 
            & 23.51\% {\scriptsize$\pm$ 0.22\%} & 
            & 0.629 {\scriptsize$\pm$ 0.009} & 
            & 0.1681 {\scriptsize$\pm$ 0.0006} & \\
            \cmidrule{1-14}
            \multirow{2}{*}{Evolutionary} 
                & AlphaEvolve           
                & \textbf{39.01\%} {\scriptsize$\pm$ 0.10\%} & 
                & \textbf{1.66} {\scriptsize$\pm$ 0.01} & 
                & \textbf{0.605} {\scriptsize$\pm$ 0.004} & 
                & 26.61\% {\scriptsize$\pm$ 0.13\%} & 
                & \textbf{0.691} {\scriptsize$\pm$ 0.009} & 
                & 0.1781 {\scriptsize$\pm$ 0.0005} & \\
                & GEPA                  
                & 38.32\% {\scriptsize$\pm$ 0.12\%} & 
                & 1.60 {\scriptsize$\pm$ 0.01} & 
                & 0.567 {\scriptsize$\pm$ 0.001} & 
                & \textbf{27.88\%} {\scriptsize$\pm$ 0.18\%} & 
                & 0.668 {\scriptsize$\pm$ 0.008} & 
                & \textbf{0.1798} {\scriptsize$\pm$ 0.0005} & \\
            \bottomrule
        \end{tabular}%
    }
    \caption{Results are reported as mean $\pm$ standard deviation over $N=10$ independent runs. \textbf{Novelty} is the dataset-level average, $\mathbb{E}_{i}[\mathrm{Novelty}_i]$, over the evaluation set. We provide detailed case studies in Appendix~\ref{app:case_study}.}
    \label{tab:main_results}
\end{table*}

\paragraph{Setup.}
We use \textsc{Qwen2.5-7B-Instruct} as the base model.
We select Layer 26 and the default steering strength $\alpha=0.1$ on a small validation set ($N=20$), following prior activation-steering and representation-engineering practice~\citep{zou2025representationengineeringtopdownapproach,turner2024steeringlanguagemodelsactivation}.
Robustness to layer and $\alpha$ sweeps is reported in Appendix~\ref{app:evorepe_ablation}.

\section{Results}
\label{sec:Experimental Results}
\subsection{CreativeBench}

As shown in Figure~\ref{fig:foundation_model}, CreativeBench poses a significant challenge even for state-of-the-art foundation models.
The strongest performing model, Gemini-3-Pro, achieves Pass@1 rates below 60\% on both subsets, reflecting the benchmark's difficulty, as it is derived from high-difficulty seeds to curb memorization and elicit creative problem solving.

\paragraph{Scaling Favors Combination over Exploration.}
We also find that \texttt{Gemini-3-Pro} achieves an improvement in \emph{combinatorial creativity}, while \emph{exploratory creativity} occurs slightly declining.
This asymmetry can be understood from a compression perspective. Training large language models can be viewed as compressing massive corpora into a finite set of parameters~\citep{delétang2024languagemodelingcompression}. Scaling expands this compression budget and increases the diversity of patterns that can be stored. Models become better at identifying deep commonalities and connecting distant domains. For example, drawing analogies between Greek literature and quantum mechanics naturally supports knowledge synthesis and recombination.

In contrast, exploratory creativity often requires moving away from dominant solution patterns, making a “0-to-1” leap into low-probability regions of the model’s prior. However, scaling that improves compression can also strengthen distributional priors, making high-likelihood, routine solutions more stable. When novelty depends on escaping these data-induced attractors, the benefits of scaling quickly diminish. Consequently, while scaling is effective for richer recombination, it may saturate for genuine exploratory innovation, which likely requires stronger search, navigation, or reasoning mechanisms beyond simple more data.

\subsection{EvoRePE}

While evolutionary algorithms effectively boost creativity, they face two critical bottlenecks. First, the process is computationally costly, as the search for novelty requires massive inference overhead (Appendix~\ref{app:evorepe_overhead}). Second, these systems are constrained by the creativity of the foundation model. In contrast, EvoRePE proposes a training-free method by injecting the creativity vector at inference time.

\paragraph{The Efficacy of the Creativity Vector.}
As shown in Table~\ref{tab:main_results}, EvoRePE yields creativity gains that are orthogonal to the evolutionary strategy. In particular, for \texttt{Qwen2.5-7B-Instruct} on CreativeBench-Combo, adding EvoRePE improves the overall creativity score from $0.174$ to $0.192$ when combined with AlphaEvolve. Notably, EvoRePE also provides consistent gains even on vanilla prompting, without requiring any evolutionary search. This indicates that the benefits of evolutionary optimization can be partially internalized into the model's activations. Our findings link evolutionary searching with latent-space steering, suggesting a promising direction of \emph{steered evolution}, where a model’s own creative trajectories facilitate a form of self-evolution in representation space.

% \section{Experimental setup}

% We evaluate a diverse range of language models, including the Qwen2.5/3 series (1.5B--72B), DeepSeek-V3, Claude-3.5-Sonnet, and multiple Gemini-2.5 variants. We compare standard zero-shot prompting against two evolutionary optimization baselines: \textbf{AlphaEvolve}~\citep{novikov2025alphaevolve} and \textbf{GEPA}~\citep{agrawal2025gepa}, using Gemini-2.5-Pro as the backend for iterative prompt mutation and feedback.

% To further enhance model performance, we adopt a training-free \textbf{representation engineering} strategy~\citep{zou2023representation, turner2023activation}. Specifically, we compute a \emph{creativity vector} by applying PCA to contrastive hidden states extracted from vanilla generations and AlphaEvolve-guided generations (Layer~26 of Qwen2.5-7B in our experiments). During inference, this vector is injected into the model’s activation space using a steering coefficient of $0.1$, biasing the model toward more creative solution patterns.

% All generated solutions are executed and verified inside an isolated sandbox environment. Semantic novelty metrics are computed using OpenAI’s \texttt{text-embedding-3-small} and \texttt{text-embedding-3-large} models.

% \input{tables/main_results.tex}
% \input{tables/base_model}

\section{Discussion and Analysis}

\begin{figure}[t]
    \centering
    \includegraphics[width=1\linewidth]{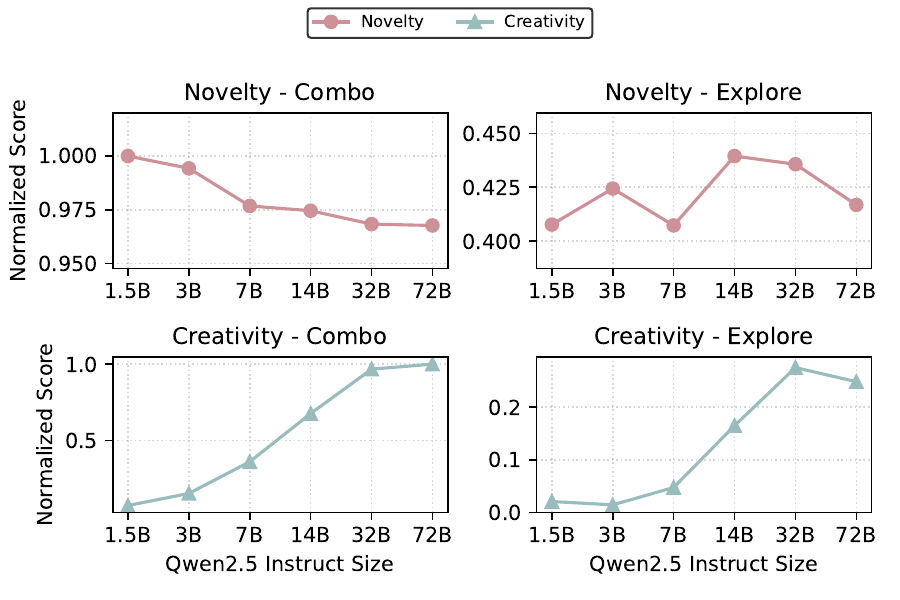}
    \caption{Scaling analysis of the Qwen2.5-Instruct model family on CreativeBench.}
    \label{fig:qwen_size}
\end{figure}

\paragraph{Convergence-by-Scaling.}
For direct comparability across tracks, we normalize Novelty and Creativity by the global maximum value computed over both Combo and Explore.
As shown in Figure~\ref{fig:qwen_size}, scaling up model size consistently improves Pass@1, while Novelty declines or plateaus, so the overall Creativity Score rises mainly due to functional gains rather than stronger divergence.
We term this \textbf{Convergence-by-Scaling}: larger models fit high-frequency training patterns more effectively, concentrating generation toward high-probability modes and yielding solutions that are more correct but also more standardized.
By contrast, smaller models exhibit higher-variance generation trajectories that deviate more from common paradigms and can yield higher novelty, typically at the cost of correctness.
Overall, Novelty and functional quality behave as largely orthogonal dimensions: scaling primarily strengthens correctness, but does not systematically increase departure from training priors.

\begin{figure}[t]
    \centering
    \includegraphics[width=1\linewidth]{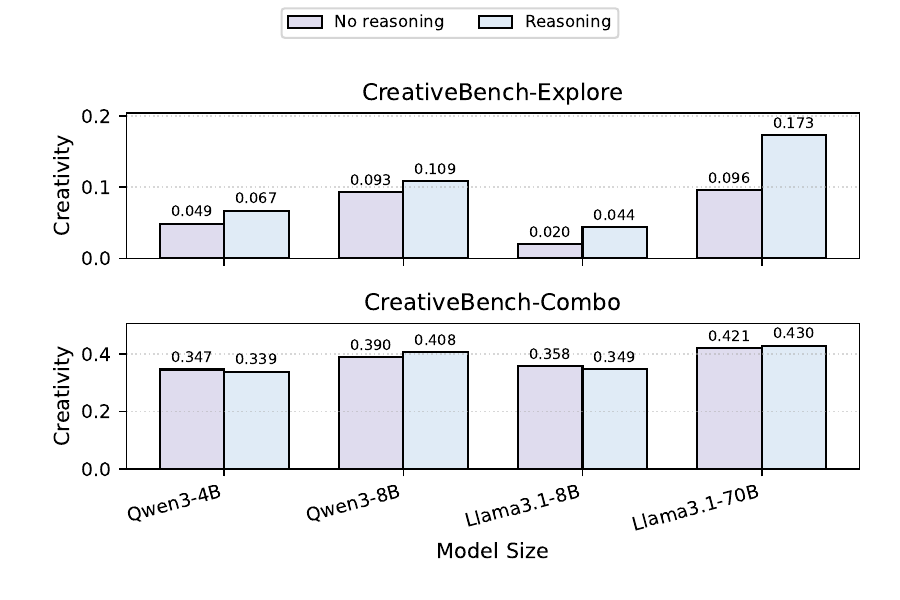}
    \caption{Impact of reasoning mode on CreativeBench.}
    \label{fig:reasoning}
\end{figure}

\paragraph{Reasoning Helps Exploratory Creativity.}
As shown in Figure~\ref{fig:reasoning}, enabling a reasoning mode has very different effects on different types of creativity. 
In combinatorial creativity tasks, reasoning provides almost no benefit, suggesting that cross-domain fusion relies more on effective knowledge retrieval and composition than on lengthy chains of thought. 
In contrast, on exploratory creativity tasks, reasoning significantly improves performance. 
When the search space is defined by constraints, a more structured ``thinking'' process helps the model find deeper alternative solutions.

\paragraph{Foundation Models as Evolutionary Operators.}
We draw an analogy between evolutionary systems and biological evolution: powerful foundation models function as effective mutation operators proposing candidate programs. In evolutionary computation, selection is driven by a fitness function that evaluates candidates; analogously, our framework instantiates the evaluation environment and computes fitness as a joint score of quality and novelty. Exploratory Creativity aligns with mutation, introducing local variations to search constrained spaces, while Combinatorial Creativity reflects recombination, merging traits from different domains into unified solutions. This suggests that further improving evolutionary systems depends on refining evolutionary operators and the fitness signals.

\paragraph{Future Work}A promising direction is to extend our evaluation framework beyond executable code to other creative domains, including storytelling, music composition, visual design, 3D artifact design, game level design, and scientific discovery. It requires (i) domain-appropriate structured representations and (ii) robust criteria for assessing both quality and novelty. While code offers execution-grounded quality signals, many domains lack standardized representations or reliable automatic evaluators. Future work could leverage domain-specific proxy metrics for quality and more principled novelty estimators, such as distance to reference sets or structural divergence over graphs and trees. Future evaluation frameworks must transcend outcome scoring to incorporate process-level signals, utilizing interaction traces to map exploration trajectories and align model behavior with human cognitive workflows. We suggest expert-in-the-loop paradigms to bridge the gap between scalability and the qualitative depth required for such assessment. A promising application area is scientific discovery, where idea generation -- such as proposing novel experimental designs -- tests the limits of generative reasoning. In this context, evaluation metrics must be expanded to rigorously quantify novelty and diversity, ensuring that generated hypotheses are both practically feasible and substantively different from established literature.

\section{Conclusion}

In this paper, we introduced \textbf{CreativeBench}, the benchmark grounded in Boden's cognitive framework to evaluate the combinatorial and exploratory creativity of evolutionary systems.
Our systematic analysis utilizing CreativeBench uncovers distinct trade-offs in modern foundation models. We identified a \emph{Convergence-by-Scaling} effect, where increasing model scale improves functional correctness but suppresses divergence. Furthermore, we found that advanced reasoning capabilities primarily benefit exploratory rather than combinatorial creativity.  To harness these findings, we proposed \textbf{EvoRePE}, a plug-and-play representation engineering strategy that steers models toward more creative behaviors by internalizing evolutionary search patterns. We believe that this work sheds light towards the exploration of machine creativity.

% Our systematic analysis reveals a critical insight: the potential of an open-ended system is fundamentally bounded by the intrinsic creativity of its underlying foundation model, which acts as the mutation operator navigating the search space.
% We observed that scaling model intelligence does not inherently guarantee diverse generative capabilities, highlighting the need for specific enhancements.
% To this end, we proposed \textbf{EvoRePE}, a plug-and-play representation engineering strategy that successfully steers models toward more creative behaviors.
% Ultimately, this work provides the essential tooling and insights to advance the field from static performance scaling to the development of genuinely self-evolving and open-ended intelligent systems.

\section*{Acknowledgements}
We thank the reviewers for their effort in improving the work. We acknowledge with thanks the discussion with Yixuan Weng, as well as the many others who have helped. This work was supported by the AI funding program of SUSTech, the National Key Research and Development Program of China (No. 2024YFB2808903), and the Tsinghua SIGS KA Cooperation Fund.

\section*{Limitations}

We acknowledge that CreativeBench has several limitations, including its language scope, training scope, and potential generator bias.

First, CreativeBench is currently instantiated in Python, whose concise syntax and mature tooling facilitate controlled analysis of novelty and functional correctness. Because the benchmark relies on an automated generation and evaluation pipeline, it is, in principle, extensible to other programming languages and paradigms through programmatic translation of code and tests.
Second, this work focuses on the \emph{evaluation and analysis} of creativity in self-evolving code generation systems, rather than on training models with CreativeBench. Due to limited computational resources, we do not conduct large-scale training or fine-tuning experiments in this study.
Finally, because CreativeBench is automatically constructed, it may inherit generator bias from the underlying LLM-based pipeline. Prior work suggests that such bias can be measured and mitigated, and under appropriate conditions, it is less likely to overturn ranking-based comparisons.

%\section*{Acknowledgments}

\bibliography{custom}

\appendix

\label{sec:appendix}

\section{Theoretical Grounding: P-Creativity vs. H-Creativity} \label{app:theoretical_grounding} To rigorously define the scope of CreativeBench, we ground our evaluation metrics in Boden's cognitive framework~\citep{Boden2004}. From a cognitive perspective, our metric is closer to ``psychological creativity'' (P-Creativity)---what the model can newly produce given its own knowledge---than to historical creativity'' (H-Creativity)---what is objectively new in human history. 
\paragraph{P-Creativity vs. H-Creativity.} P-Creativity refers to the generation of ideas that are novel to the individual agent, regardless of whether others have had the idea before. In contrast, H-Creativity requires the idea to be novel to the entire history of humanity. Defining and evaluating H-Creativity in large language models (LLMs) remains challenging. 
Although these models are trained on Internet-scale pre-training corpora that, to some extent, compress and reflect large portions of human history, their training is inherently static and cannot fully capture the most recent developments in human knowledge. 
As a result, the notion and evaluation of H-Creativity in LLMs are subject to fundamental ambiguities.
\paragraph{Novelty as Deviation from Priors.} Consequently, CreativeBench focuses on P-Creativity. We operationalize this by measuring \textit{Novelty} as the distance between the model's generated solution and a ``standard'' baseline (representing the model's default behavior or high-probability path). If a model can generate a correct solution that significantly deviates from its own statistical priors (the most likely path) or the constituent source components, it demonstrates P-Creativity by traversing new regions of its latent solution space. This approach allows for a quantitative assessment of the system's generative flexibility without the ambiguity of verifying historical uniqueness.

\section{Experimental Setup}
\label{sec:setup}
\subsection{LLM-as-a-Judge Evaluation Prompt}
\label{appendix:judge-spec}
This appendix provides the full prompt used for the LLM-as-a-Judge component described in Table~\ref{tab:JudgePrompt}. The Judge Prompt contains the instruction set used by the LLM-as-a-Judge component to verify constraint compliance in CreativeBench-Explore. It evaluates whether a generated solution adheres to blocked techniques and demonstrates genuine exploratory creativity.
Based on human-labeled set, the \textit{Constraint Compliance} judge achieves 94\% precision and 91\% recall.
\subsection{Implementation details.}
For all experiments and methods, we fix the decoding configuration for fair comparison: temperature $=0.1$, top-$p=1.0$, and top-$k=0$, which reduces to greedy decoding in our inference stack. We run three independent trials with seeds $\{42,43,44\}$ and report the mean across runs unless stated otherwise.

\section{Additional Benchmark Comparison}
\label{app:benchmark_comparison}

Table~\ref{tab:comparisons} summarizes representative code generation benchmarks and contrasts them with \textbf{CreativeBench} along several axes.
Compared to conventional correctness-only evaluations (e.g., Pass@$k$), we explicitly highlight whether a benchmark targets \emph{creative behaviors} (exploratory or combinatorial), and whether it supports a fully automated (human-free) construction pipeline.
We also report coarse task difficulty (as a qualitative indicator), the number of covered domains, and the average length of problem descriptions (\textbf{Len}).
Note that the difficulty stars are intended as a high-level proxy rather than a standardized measure, and \textbf{Len} depends on the tokenization/counting protocol used in each benchmark.

\section{Data Construction Pipeline}
\label{app:algorithms}
We construct CreativeBench using a fully automated pipeline. 
CreativeBench spans 14 practical programming domains (tags), ranging from core language fundamentals and data structures to web, systems, and ML-oriented tasks, including Algorithms \& Problem Solving, Concurrency \& Async Programming, Data Structures \& Collections, Language Fundamentals, Functions \& Modules, Web Development \& Frameworks, Systems Programming \& Low-level Development, Network Programming \& Communication, Data Science \& Analytics, File \& I/O Operations, Machine Learning \& AI, Database Operations \& Persistence, Error Handling \& Debugging, and Others (where categories with less than 2\% representation are merged).
To ensure both high difficulty and strict quality control without manual curation, we design two distinct generation paradigms: a \emph{reverse-engineering} pipeline for combinatorial tasks and a \emph{self-play} pipeline for exploratory tasks.

\begin{algorithm}[!h]
\caption{Construction Pipeline for CreativeBench-Combo}
\label{alg:combo_pipeline}
\begin{algorithmic}[1]
\Require Seed code components $\mathcal{D}$, Foundation Model $\mathcal{M}$, Sandbox $\mathcal{E}$
\Ensure Combinatorial Dataset $\mathcal{S}_{combo}$
\State $\mathcal{S}_{combo} \leftarrow \emptyset$
\For{each iteration $i = 1$ to $N$}
    \State \textcolor{gray}{\emph{// Step 1: Solution Fusion}}
    \State $S_{ref} \leftarrow \mathcal{M}.\text{fuse\_components}(\mathcal{D})$ 
    \If{$\mathcal{E}.\text{execute}(S_{ref}) \neq \text{Success}$} 
        \State \textbf{continue}
    \EndIf
    
    \State \textcolor{gray}{\emph{// Step 2: Test Function Generation}}
    \State $I \leftarrow \mathcal{M}.\text{generate\_inputs}(S_{ref})$
    \State $O \leftarrow \mathcal{E}.\text{get\_outputs}(S_{ref}, I)$ \Comment{Get ground truth via sandbox}
    \State $T_{func} \leftarrow \text{construct\_test\_function}(I, O)$
    
    \State \textcolor{gray}{\emph{// Step 3: Problem Synthesis (Reverse Engineering)}}
    \State $P_{desc} \leftarrow \mathcal{M}.\text{synthesize\_problem}(S_{ref}, T_{func})$
    
    \State \textcolor{gray}{\emph{// Step 4: Filtering}}
    \If{$\text{Filter}(P_{desc}, T_{func})$ is \textbf{True}}
        \State $\mathcal{S}_{combo} \leftarrow \mathcal{S}_{combo} \cup \{(P_{desc}, S_{ref}, T_{func})\}$
    \EndIf
\EndFor
\end{algorithmic}
\end{algorithm}

\begin{algorithm}[!h]
\caption{Construction Pipeline for CreativeBench-Explore}
\label{alg:explore_pipeline}
\begin{algorithmic}[1]
\Require Seed Task $T_{seed} = (P, S_{base})$, Generator $\mathcal{G}$, Solver $\mathcal{M}$, Sandbox $\mathcal{E}$, Judge $\mathcal{J}$
\Ensure Exploratory Dataset $\mathcal{S}_{explore}$
\State $\mathcal{S}_{explore} \leftarrow \emptyset$
\For{each task $(P, S_{base}) \in T_{seed}$}
    \State \textcolor{gray}{\emph{// Step 1: Targeted Constraint Injection}}
    \State $\mathcal{C} \leftarrow \mathcal{G}.\text{analyze\_and\_constrain}(S_{base})$ \Comment{E.g., forbid sort(), max()}
    
    \State \textcolor{gray}{\emph{// Step 2: Reference-Guided Refinement (Self-Play)}}
    \State $S_{curr} \leftarrow S_{base}$
    \State $Solved \leftarrow \textbf{False}$
    
    \While{\textbf{not} $Solved$ \textbf{and} steps $<$ MaxSteps}
        \State $S_{new} \leftarrow \mathcal{M}.\text{refine}(S_{curr}, \mathcal{C}, \text{Feedback})$
        
        \State $v_{exec} \leftarrow \mathcal{E}.\text{execute}(S_{new})$ \Comment{Check correctness}
        \State $v_{const} \leftarrow \mathcal{J}.\text{check\_constraint}(S_{new}, \mathcal{C})$ \Comment{Check novelty}
        
        \If{$v_{exec} \land v_{const}$}
            \State $\mathcal{S}_{explore} \leftarrow \mathcal{S}_{explore} \cup \{(P, \mathcal{C}, S_{new})\}$
            \State $Solved \leftarrow \textbf{True}$
        \Else
            \State $\text{Feedback} \leftarrow \text{generate\_feedback}(v_{exec}, v_{const})$
            \State $S_{curr} \leftarrow S_{new}$
        \EndIf
    \EndWhile
\EndFor
\end{algorithmic}
\end{algorithm}

\subsection{Data Filtering}
\label{sec:Data_Filtering}
\subsubsection{Consistency Specifications}

The Judge evaluates each sample according to the following criteria:

1. Signature Consistency: Function names, class names, and parameters must match the problem description.

2. Randomness Handling: Test cases must not rely on non-reproducible randomness (e.g., unset random seeds).

3. Objective Alignment: Test logic must faithfully verify the intended objective of the problem rather than unrelated behaviors.

4. Numerical Precision: Floating-point operations must properly handle rounding/epsilon issues.

5. Exception Safety: Broad \texttt{try--except} blocks must not suppress assertion failures or mask genuine errors.

6. Requirement Hallucination: Tests must not enforce constraints that are absent from the problem description.

7. Test Completeness: Essential corner cases and boundary conditions must be covered.

\subsubsection{Quality Audit Prompt}
As shown in Table~\ref{tab:QualityPrompt}. The Quality Audit Prompt (Table 4) provides the full specification used to assess the correctness, clarity, and robustness of each dataset record. It ensures that problem statements, reference solutions, and test suites are well-aligned and resistant to trivial or unintended shortcuts.

\subsubsection{Human Evaluation}
To conduct a human study, we employ three master's students in computer science and compensate them at \$33/hour (4 hours/day for 15 days). All annotators have basic familiarity with large language models and Python programming. The authors double-check annotations daily and provide feedback. On this human-labeled set, the overall validity rate is 89.1\%, and the automated creativity ranking is highly consistent with expert judgments (Spearman's $\rho=0.78$).

\noindent The experts evaluate each sample using the following criteria:
\begin{enumerate}
    \item \textbf{Signature Consistency:} Function names, class names, and parameters must match the problem description.
    \item \textbf{Randomness Handling:} Test cases must not rely on non-reproducible randomness (e.g., unset random seeds).
    \item \textbf{Objective Alignment:} Test logic must faithfully verify the intended objective of the problem rather than unrelated behaviors.
    \item \textbf{Numerical Precision:} Floating-point operations must properly handle rounding/epsilon issues.
    \item \textbf{Exception Safety:} Broad \texttt{try--except} blocks must not suppress assertion failures or mask genuine errors.
    \item \textbf{Requirement Hallucination:} Tests must not enforce constraints that are absent from the problem description.
    \item \textbf{Test Completeness:} Essential corner cases and boundary conditions must be covered.
\end{enumerate}

\section{Additional Experimental Details}
\label{app:extra_details}

\subsection{Paired Significance Tests}
\label{app:significance}

We conduct paired significance tests over $N=10$ matched random seeds. For each metric, we compute per-seed differences
$d_i = \textsc{EvoRePE} - \text{baseline}$ and run a two-sided paired $t$-test ($\mathrm{df}=9$).
\textsc{EvoRePE} yields significant gains on the main metrics:
Combo Pass@1 $+0.80$pp (95\% CI $[0.68, 0.92]$, $p=2\times 10^{-9}$),
Combo Creativity $+0.020$ (95\% CI $[0.016, 0.024]$, $p=5\times 10^{-10}$),
Explore Pass@1 $+0.60$pp (95\% CI $[0.45, 0.75]$, $p=3\times 10^{-6}$),
and Explore Creativity $+0.0012$ (95\% CI $[0.0008, 0.0016]$, $p=1\times 10^{-7}$).
Overall, paired-seed tests support gains beyond seed variance.

\subsection{Wall-Clock Cost of the Full Pipeline}
\label{app:runtime}

We report wall-clock costs from logged runs on the target splits.
For Combo (dataset\_items$=1308$, 9 runs), the median time is 3.19 hours (P25/P75: 0.92/4.47h).
For Exploration (total\_items/prompts$=551$, 29 runs), the median time is 8.40 hours (P25/P75: 2.42/9.20h).
A representative end-to-end run combining Combo and Exploration takes $\sim$8.78 hours.

\subsection{Baselines: AlphaEvolve (OpenEvolve) and GEPA Hyperparameter Settings}
\label{app:baseline_hparams}

\paragraph{AlphaEvolve (OpenEvolve implementation).}
We run the AlphaEvolve-style evolutionary coding baseline using the open-source \texttt{OpenEvolve} framework. Unless otherwise specified, we adopt the framework's default ``balanced'' configuration (dataclass defaults) for the evolutionary database and selection, together with the default evaluator and LLM settings. We set the evolution budget to \texttt{max\_iterations}=10000 with checkpoints every \texttt{checkpoint\_interval}=100 iterations. The population and archive sizes are \texttt{population\_size}=1000 and \texttt{archive\_size}=100. We use island-model evolution with \texttt{num\_islands}=5, migrating every \texttt{migration\_interval}=50 generations at \texttt{migration\_rate}=0.1. Selection uses \texttt{elite\_selection\_ratio}=0.1 with \texttt{exploration\_ratio}=0.2 and \texttt{exploitation\_ratio}=0.7. We enable OpenEvolve's internal deduplication/novelty filter with \texttt{similarity\_threshold}=0.99 (cosine similarity over embeddings) and use diff-based evolution (\texttt{diff\_based\_evolution}=true) with \texttt{max\_code\_length}=10000.

\paragraph{GEPA.}
We run GEPA using the official \texttt{gepa.optimize()} API with its default candidate selection, frontier tracking, and reflective mutation settings, and we specify an explicit metric-call budget of \texttt{max\_metric\_calls}=150. We set \texttt{candidate\_selection\_strategy}=\texttt{"pareto"} and \texttt{frontier\_type}=\texttt{"instance"}. For batching, we use \texttt{batch\_sampler}=\texttt{"epoch\_shuffled"} with \texttt{reflection\_minibatch\_size}=3, and we enable \texttt{skip\_perfect\_score}=true with \texttt{perfect\_score}=1.0. Components are updated in a round-robin manner via \texttt{module\_selector}=\texttt{"round\_robin"}. We keep merging disabled (\texttt{use\_merge}=false), disable evaluation caching (\texttt{cache\_evaluation}=false), and set \texttt{seed}=0 for reproducibility.

% =========================
% Add this to Appendix
% =========================
\section{Additional Robustness Checks for the Novelty}
\label{app:novelty_robustness}

A potential concern is that the character-level 4-gram novelty term may be overly sensitive to superficial edits
(e.g., identifier renaming, formatting, or comment changes), and that an unnormalized equal-weight sum could introduce
length/scale artifacts without adversarial robustness checks.
In response, we provide additional stress tests to characterize how each novelty component behaves under non-semantic
code perturbations and length variation.

\paragraph{Metric components and bounded scale.}
Our novelty score is intentionally hybrid, combining (i) a semantic embedding distance and (ii) a lexical char-4gram distance.
Both components are ratio-based and bounded, which reduces uncontrolled scale drift:
the embedding term uses cosine distance ($d_{\text{embed}} = 1-\cos(\cdot,\cdot)$), and the n-gram term uses a Jaccard
distance over character 4-grams ($d_{\text{ngram}} \in [0,1]$).
We do not claim strict invariance to length or surface edits; instead, we empirically quantify residual effects below.

\paragraph{Canonicalization to reduce surface-form sensitivity.}
To mitigate purely formatting-based rewrites, before computing the char-4gram term we preprocess generated solutions
to canonicalize superficial surface forms, including normalizing whitespace/indentation and stripping comment-only changes.
This reduces sensitivity to adversarial reformatting while preserving sensitivity to genuine lexical rewrites.

\subsection{Embedding robustness under superficial edits}
\label{app:embed_robustness}

We first isolate the semantic stability of the embedding-based signal by measuring only
$d_{\text{embed}}$ (cosine distance) on functionally equivalent code under controlled, semantics-preserving edits.
We use CodeXEmbed as the embedding model, producing 2304-dimensional L2-normalized embeddings.

\begin{table*}[t]
\small
\centering
\setlength{\tabcolsep}{4pt}
\begin{tabular}{lcc}
\toprule
\textbf{Perturbation type (embedding-only)} & $d_{\text{embed}}$ & \textbf{Share of full range} \\
\midrule
Single-variable rename (light rename) & 0.0008 & 0.04\% \\
Formatting-only edits & 0.0045 & 0.22\% \\
Short comment addition & 0.0075 & 0.38\% \\
Moderate comment-only length increase (184 $\rightarrow$ 801 chars) & 0.0333 & 1.67\% \\
Chunk-length setting shift (32{,}768 vs 128{,}000) & 0.0040 & 0.20\% \\
\bottomrule
\end{tabular}
\caption{Controlled sanity checks for the embedding-based novelty signal under non-semantic edits.}
\label{tab:embed_sanity}
\end{table*}

\paragraph{Contextualizing magnitudes via cross-model baselines.}
To contextualize the above values, we compute cross-model embedding distances between solutions on the same tasks
(120 shared combo problems; GPT-4.1-nano, Gemini-2.5-pro, Qwen2.5-Coder-1.5B).

\begin{table}[t]
\small
\centering
\setlength{\tabcolsep}{6pt}
\begin{tabular}{lc}
\toprule
\textbf{Statistic} & \textbf{Value} \\
\midrule
Mean & 0.0813 \\
Median & 0.0728 \\
75th percentile & 0.1000 \\
\bottomrule
\end{tabular}
\caption{Cross-model $d_{\text{embed}}$ baseline on the same problems (context for scale).}
\label{tab:embed_cross_model}
\end{table}

Relative to the cross-model mean baseline (0.0813), typical superficial edits are substantially smaller:

\begin{table*}[t]
\small
\centering
\setlength{\tabcolsep}{4pt}
\begin{tabular}{lcc}
\toprule
\textbf{Perturbation type} & $d_{\text{embed}}$ & \textbf{Ratio to cross-model mean} \\
\midrule
Single-variable rename & 0.0008 & 1.0\% \\
Formatting-only edits & 0.0045 & 5.5\% \\
Chunk-length setting shift & 0.0040 & 4.9\% \\
Short comment addition & 0.0075 & 9.2\% \\
\bottomrule
\end{tabular}
\caption{Embedding robustness: superficial edits are much smaller than cross-model differences.}
\label{tab:embed_ratio}
\end{table*}

\paragraph{Takeaway.}
Under typical non-semantic edits and moderate length variation, the embedding-based novelty signal remains stable
(nearly zero or small shifts), while cross-model comparisons yield substantially larger distances.

\subsection{Role and robustness of the character-level 4-gram term}
\label{app:ngram_robustness}

\paragraph{Complementarity of lexical vs.\ semantic signals.}
Our novelty metric is intentionally hybrid: embeddings capture semantic-level deviation, while the char-4gram term acts as
a lexical novelty regularizer and is expected to respond to genuine surface-level rewrites.
We nevertheless stress-test its sensitivity to superficial edits and length mismatch.

\paragraph{Length sensitivity at scale.}
Over 54,940 source--combo pairs, the correlation between $d_{\text{ngram}}$ and length mismatch is weak:
Pearson($d_{\text{ngram}}$, length ratio) $=0.1168$, and Pearson($d_{\text{ngram}}$, absolute length difference) $=0.1052$.

\begin{table}[t]
\small
\centering
\setlength{\tabcolsep}{6pt}
\begin{tabular}{lc}
\toprule
\textbf{Correlation test (dataset-level)} & \textbf{Pearson $r$} \\
\midrule
$d_{\text{ngram}}$ vs.\ length ratio & 0.1168 \\
$d_{\text{ngram}}$ vs.\ absolute length difference & 0.1052 \\
\bottomrule
\end{tabular}
\caption{Weak correlation between char-4gram distance and length mismatch at the dataset level.}
\label{tab:ngram_length_corr}
\end{table}

\paragraph{Controlled perturbations (functionally equivalent code).}
After canonicalization (whitespace/indent normalization; stripping comment-only changes), $d_{\text{ngram}}$ changes only
slightly under typical non-semantic edits, and increases more noticeably only under substantial comment-only length growth:

\begin{table*}[t]
\small
\centering
\setlength{\tabcolsep}{4pt}
\begin{tabular}{lc}
\toprule
\textbf{Perturbation type (canonicalized n-gram)} & $d_{\text{ngram}}$ \\
\midrule
Single light rename & 0.0149 \\
Mild formatting changes & 0.0160 \\
Short comment addition & 0.0176 \\
Substantial comment-only length increase (1104 $\rightarrow$ 2302 chars) & 0.0621 \\
\bottomrule
\end{tabular}
\caption{Controlled sanity checks for the char-4gram term under non-semantic edits.}
\label{tab:ngram_sanity}
\end{table*}

\paragraph{Adversarial reformatting and ablations.}
As expected for a lexical metric, aggressive whitespace-only reformatting (e.g., extreme indentation/tab changes) can
produce larger shifts. For transparency, we also ran an ablation without canonicalization (raw n-gram) as a worst-case
stress test to expose maximal surface-form sensitivity.

\paragraph{Normalization ablation (ranking stability).}
Model-level conclusions are stable after normalization: the Spearman correlation between original and normalized creativity
rankings is 0.9989, with a maximum rank shift of 1.

\paragraph{Overall takeaway.}
We do not claim the char-4gram term is invariant to superficial edits. Our added analyses show that (i) sensitivity is bounded,
(ii) length mismatch correlates only weakly with $d_{\text{ngram}}$ at scale, and (iii) model-level conclusions remain unchanged
under normalization/ablation, while the embedding-based signal remains highly stable under typical non-semantic edits and moderate
length variation.

\section{Case Study}
\label{app:case_study}
\subsection{Case 1: Algorithmic Search Without Binary Search}
\label{app:case_study_hamburgers}

To demonstrate that \textsc{CreativeBench} constraints elicit \emph{algorithmic} creativity beyond surface-level syntactic rewrites, we present a detailed analysis of the \textit{Maximum Hamburgers You Can Make} task (Problem 126) from the \textit{Exploratory} subset. This case highlights how a single targeted constraint (forbidding binary search) compels the model to devise a qualitatively different search strategy under a monotone feasibility structure.

\subsubsection{Task Specification and Constraints}
The task asks for the maximum number of hamburgers producible given a recipe, available ingredients, per-unit prices, and a total budget. While the canonical solution relies on the monotonicity of the cost function and employs binary search, we explicitly forbid this dominant idiom, forcing the model into alternative search procedures~\citep{xu2025alignmentefficienttoolcalling,xu2025reducingtoolhallucinationreliability,zhou2025mementofinetuningllmagents, xiao2026not,huang2025critictool,an2025amo, cao2026dpwriter}.

% ---------------------------------------------------------
% 题目描述框 (建议保留为LaTeX文本，清晰可搜)
% ---------------------------------------------------------
\begin{figure}[!th]
    \centering
    \begin{tabular}{|p{0.96\linewidth}|}
        \hline
        \textbf{Problem 126: Maximum Hamburgers You Can Make} \\
        \hline

        \textbf{Problem Description:} \\
        Write a Python function \texttt{max\_hamburgers(recipe, available, price, budget)} that:
        \begin{itemize}
            \item Takes a recipe string over \{B, S, C\} (Bread, Sausage, Cheese).
            \item Takes \texttt{available} and \texttt{price} as length-3 integer lists in the order [B, S, C].
            \item Returns the maximum integer $n$ such that one can produce $n$ hamburgers by using available ingredients and buying additional units within \texttt{budget}.
            \item If \texttt{recipe} is empty, returns a very large number (e.g., $10^{18}$) per specification.
        \end{itemize}

        \textbf{Negative Constraints ($\mathcal{C}$):} \\
        The following pattern is \textbf{strictly forbidden}:
        \begin{enumerate}
            \item \textbf{No Binary Search:} Do not use binary search (no interval-halving logic over $n$).
        \end{enumerate}
        \\
        \hline
    \end{tabular}
    \caption{The full specification for Problem 126. By forbidding binary search---the dominant idiom for monotone feasibility---the task forces models to synthesize alternative search strategies.}
    \label{fig:hamburger_task_spec}
\end{figure}

\subsubsection{Analysis of Creative Restructuring}
%The constrained solution achieves a substantial \textit{Creativity Gap} ($0.3608$), not by modifying syntax, but by re-architecting the \emph{search procedure}. Concretely, it replaces interval halving with a two-stage strategy: (i) derive a conservative analytic upper bound, and (ii) perform a coarse-to-fine decremental search over decreasing step sizes.

\paragraph{1. Monotone Cost Modeling.}
Both baseline and constrained solutions rely on the monotone structure of the purchasing cost:
\[
\mathrm{cost}(n)=\sum_{i\in\{B,S,C\}} \max(0,\; n\cdot \mathrm{need}_i-\mathrm{avail}_i)\cdot \mathrm{price}_i,
\]
which is non-decreasing in $n$. The baseline exploits this monotonicity via binary search. Under our constraint, the model must preserve the same semantic invariant while abandoning the canonical optimizer.

\paragraph{2. Conservative Upper Bounding as a Search Pivot.}
The constrained solution computes the ``free'' production limit (without buying),
\[
n_{\text{free}}=\min_{i:\,\mathrm{need}_i>0}\left\lfloor \frac{\mathrm{avail}_i}{\mathrm{need}_i}\right\rfloor,
\]
and then derives a loose but safe upper bound by assuming each additional hamburger requires purchasing \emph{all} required ingredients:
\[
n_{\text{ub}} \;=\; n_{\text{free}} \;+\; \left\lfloor \frac{\mathrm{budget}}{\sum_i \mathrm{need}_i\cdot \mathrm{price}_i}\right\rfloor.
\]
This bound does not need to be tight; its role is to locate a region near feasibility while respecting the constraint.

\paragraph{3. Algorithmic Morphing: Coarse-to-Fine Step-Down Search.}
Instead of halving an interval, the model performs a multi-resolution descent:
\[
n \leftarrow n_{\text{ub}},\\
\text{for  } s\in\{10^{12},10^{11},\dots,10,1\}: \;\; 
\]
while $\mathrm{cost}$ $(n)>\mathrm{budget}:\;\; n\leftarrow n-s$. This procedure can be interpreted as a digit-wise refinement in base-10: large steps quickly correct order-of-magnitude errors, while smaller steps finalize the exact maximum feasible $n$. Crucially, it avoids the structural signature of binary search (midpoint selection and repeated halving), yet still leverages monotonicity to guarantee convergence to a feasible boundary.

%\paragraph{Code (Baseline vs. Constrained)}
%We include the complete code for both baseline and constrained solutions below.

% ---------------------------------------------------------
% 代码占位符：后续填入完整代码（Baseline + Constrained）
% ---------------------------------------------------------
\begin{figure*}[!th]
    \centering
    % -----------------------------
    % Baseline (left) vs Constrained (right)
    % -----------------------------
    \begin{minipage}[t]{0.49\linewidth}
        \centering
        \captionof{lstlisting}{\textbf{Problem 126 (Baseline)}: Standard solution using binary search.}
        \label{lst:hamburger_baseline}
\begin{lstlisting}[language=Python]
def count_ingredients(recipe_str):
    """
    Counts the number of each ingredient in the recipe string.
    Returns a list in the order [Bread, Sausage, Cheese].
    """
    return [recipe_str.count('B'), recipe_str.count('S'), recipe_str.count('C')]

def max_hamburgers(recipe, available, price, budget):
    """
    Calculates the maximum number of hamburgers that can be made
    given the recipe, available ingredients, their prices, and the budget.
    """
    # Using count_ingredients as a creative 'helper' to parse the recipe
    need = count_ingredients(recipe)

    # If recipe is empty, infinite hamburgers can be made
    if sum(need) == 0:
        return int(1e18)

    # Helper to compute cost for making 'n' hamburgers
    def cost_for(n):
        total = 0
        for i in range(3):
            required = max(0, need[i] * n - available[i])
            total += required * price[i]
        return total

    # Binary search for the answer
    lo, hi = 0, 10**13
    while lo < hi:
        mid = (lo + hi + 1) // 2
        if cost_for(mid) <= budget:
            lo = mid
        else:
            hi = mid - 1
    return lo
\end{lstlisting}
    \end{minipage}
    \hfill
    \begin{minipage}[t]{0.49\linewidth}
        \centering
        \captionof{lstlisting}{\textbf{Problem 126 (Constrained)}: Avoids binary search via step-wise descent.}
        \label{lst:hamburger_constrained}
\begin{lstlisting}[language=Python]
def max_hamburgers(recipe, available, price, budget):
    # Map ingredient indices
    idx = {'B': 0, 'S': 1, 'C': 2}
    # Count needed per hamburger
    from collections import Counter
    need = [0, 0, 0]
    for k, v in Counter(recipe).items():
        need[idx[k]] = v

    # Special case: empty recipe
    if sum(need) == 0:
        return int(1e18)

    # If need[i] == 0, set available[i] and price[i] to 0 so they don't affect calcs
    # Compute max hamburgers we can make WITHOUT buying
    possible = float('inf')
    for i in range(3):
        if need[i]:
            possible = min(possible, available[i] // need[i])
    if possible == float('inf'):
        possible = 0

    # Compute per-hamburger cost if we have to buy all ingredients
    per_hamburger_cost = sum(need[i] * price[i] for i in range(3))
    # Loose but safe upper bound
    max_by_budget = possible + budget // (per_hamburger_cost if per_hamburger_cost else 1)

    def cost_for(n):
        buy = [max(0, need[i] * n - available[i]) for i in range(3)]
        return sum(buy[i] * price[i] for i in range(3))

    if cost_for(max_by_budget) <= budget:
        return max_by_budget

    n = max_by_budget
    for step in [
        int(1e12), int(1e11), int(1e10), int(1e9), int(1e8), int(1e7),
        int(1e6), int(1e5), int(1e4), int(1e3), int(1e2), int(1e1), 1
    ]:
        while n >= possible and cost_for(n) > budget:
            n -= step
    return n
\end{lstlisting}
    \end{minipage}

    \caption{Full code listing for Problem 126. Left: baseline solution using binary search. Right: constrained solution that replaces binary search with a step-wise descent heuristic under the ``no binary search'' constraint.}
    \label{fig:hamburger_code}
\end{figure*}

\subsubsection{Discussion}
This case study illustrates that \textsc{CreativeBench} constraints can elicit \emph{algorithmic restructuring} rather than mere syntactic variation. The constrained solution exhibits a clear departure from the dominant binary-search template: it constructs an analytic upper bound and executes a coarse-to-fine step-down procedure to locate the maximum feasible output under a monotone cost model. This behavior is precisely the kind of low-probability, structure-altering adaptation our \textit{Exploratory} tasks are designed to measure.

\subsection{Case 2: Algorithmic and Syntactic Restructuring}

To demonstrate the granularity of creativity elicited by \textsc{CreativeBench}, we present a detailed analysis of the \textit{Temperature Conversion Table} task (Problem 192) from the \textit{Exploratory} subset. This case highlights how fine-grained constraints compel the model to perform significant \textbf{syntactic restructuring} and \textbf{mathematical decomposition}.

\subsubsection{Task Specification and Constraints}
The core task is straightforward: generate a Fahrenheit-to-Celsius table. However, as shown in Figure~\ref{fig:temp_task_spec}, we impose a set of ``scorched-earth'' negative constraints designed to block all idiomatic Python solutions. By forbidding loops, standard formulas, and string formatting, we force the model into a low-probability search space.

% ---------------------------------------------------------
% 题目描述框 (建议保留为LaTeX文本，清晰可搜)
% ---------------------------------------------------------
\begin{figure}[!th]
    \centering
    % 使用 tabular 创建一个占满宽度的带边框盒子
    \begin{tabular}{|p{0.96\linewidth}|}
        \hline
        % 标题部分
        \textbf{Problem 192: Temperature Conversion Table} \\
        \hline
        
        \textbf{Problem Description:} \\
        Write a Python function \texttt{print\_temperature\_table(start, end)} that:
        \begin{itemize}
            \item Takes two integers \texttt{start} and \texttt{end}.
            \item If \texttt{start > end}, returns the string ``Invalid.''.
            \item Otherwise, prints a table with columns ``Fahrenheit'' and ``Celsius'' (rounded to 2 decimal places).
            \item Returns \texttt{None} upon success.
        \end{itemize}
        
        \textbf{Negative Constraints ($\mathcal{C}$):} \\
        The following patterns are \textbf{strictly forbidden}:
        \begin{enumerate}
            \item \textbf{No Iteration Primitives:} Do not use \texttt{for} loops, \texttt{while} loops, or \texttt{range()}.
            \item \textbf{No Standard Formula:} Do not use the arithmetic formula $5 \times (F - 32) / 9$ or any direct equivalent (e.g., $0.555\dots$, $1.8$).
            \item \textbf{No Syntactic Sugar:} Do not use formatted printing (\texttt{f-string}, \texttt{.format()}, \texttt{\%}).
            \item \textbf{No Early Return:} Do not use early returns for input validation.
        \end{enumerate} 
        \\
        \hline
    \end{tabular}
    \caption{The full specification for Problem 192. The combination of a simple functional goal with severe syntactic restrictions forces the model to abandon standard programming paradigms.}
    \label{fig:temp_task_spec}
\end{figure}

\subsubsection{Analysis of Creative Restructuring}
The model (Qwen2.5-72B-Instruct) successfully navigated these constraints, achieving a high \textit{Creativity Gap} of $0.3999$. The generated solution (shown in Figure~\ref{fig:case_study_comparison}) exhibits innovation across three dimensions:

\paragraph{1. Mathematical Semantic Decomposition.} 
To circumvent the ban on the constant $\frac{5}{9}$, the model did not merely use a different approximation. Instead, it demonstrated deep semantic understanding by analytically decomposing the fraction:
\[
\frac{5}{9} = \frac{1}{3} +  \frac{2}{9}
\]
As seen in the solution code, the model implements this using the \texttt{decimal} module to ensure precision, adhering to the letter of the constraint while preserving mathematical exactness.

\paragraph{2. Structural Morphing: Recursion over Iteration.}
Deprived of \texttt{for} loops, the model fundamentally altered the code's Abstract Syntax Tree (AST). It implemented a recursive helper function \texttt{\_print\_row\_recursive} to traverse the temperature range. This shift from an iterative to a recursive paradigm represents a significant departure from standard training distributions for simple I/O tasks.

\paragraph{3. Primitive I/O Management.}
Without high-level formatting tools like f-strings, the model reverted to manipulating low-level system streams. It utilized \texttt{sys.stdout.write} combined with manual string alignment methods (\texttt{.ljust}, \texttt{.rjust}). This demonstrates the model's versatility in degrading gracefully from high-level abstractions to primitive operations while maintaining correctness.

% ---------------------------------------------------------
% Replace the screenshot-based figure with verbatim code.
% ---------------------------------------------------------
\begin{figure*}[!th]
\centering
\begin{minipage}[t]{0.49\linewidth}
\caption*{\textbf{(a) Baseline Solution}\\
Standard Python approach using \texttt{for}-loops and f-strings.}
\begin{lstlisting}[basicstyle=\ttfamily\scriptsize]
def fahrenheit_to_celsius(fahrenheit):
    """
    Converts a temperature from Fahrenheit to Celsius using the standard formula.
    C = (F - 32) * 5 / 9
    """
    return (fahrenheit - 32) * 5 / 9

def print_temperature_table(start, end):
    """
    Generates and prints a temperature conversion table from Fahrenheit to Celsius
    for a given range of Fahrenheit temperatures.

    Args:
        start (int): The starting Fahrenheit temperature (inclusive).
        end (int): The ending Fahrenheit temperature (inclusive).

    Returns:
        str: "Invalid." if start > end.
        None: If the table is printed successfully.
    """
    if start > end:
        return "Invalid."

    # Define headers for the table
    header_f = "Fahrenheit"
    header_c = "Celsius"

    # Store all Fahrenheit and calculated Celsius values first.
    # This allows for dynamic calculation of column widths for perfect alignment.
    table_data = []
    for f_temp in range(start, end + 1):
        c_temp = fahrenheit_to_celsius(f_temp)
        table_data.append((f_temp, c_temp))

    # Determine the maximum width required for each column,
    # considering both headers and all data points (including formatting for Celsius).
    max_f_width = len(header_f)
    max_c_width = len(header_c)

    for f_val, c_val in table_data:
        max_f_width = max(max_f_width, len(str(f_val)))
        # Use f-string formatting to get the length of the rounded Celsius value
        max_c_width = max(max_c_width, len(f"{c_val:.2f}"))

    # Assemble the table lines into a list of strings
    table_lines = []

    # Add the header row: Fahrenheit left-aligned, Celsius right-aligned for numbers
    table_lines.append(f"{header_f:<{max_f_width}}   {header_c:>{max_c_width}}")

    # Add a separator line for visual clarity
    table_lines.append(f"{'-'*max_f_width}   {'-'*max_c_width}")

    # Add data rows: Fahrenheit values left-aligned, Celsius values right-aligned and formatted
    for f_val, c_val in table_data:
        table_lines.append(f"{f_val:<{max_f_width}}   {c_val:>{max_c_width}.2f}")

    # Print the entire table by joining all lines with a newline character
    print("\n".join(table_lines))

    return None
\end{lstlisting}
\end{minipage}
\hfill
\begin{minipage}[t]{0.49\linewidth}
\caption*{\textbf{(b) Constrained Solution (Ours)}\\
Replaces loops with \textbf{recursion} and uses raw I/O to bypass formatting constraints.}
\begin{lstlisting}[basicstyle=\ttfamily\tiny]
import sys
from decimal import Decimal, getcontext
# This helper function is required by the problem statement's signature constraint.
# It calculates Celsius from Fahrenheit using a non-standard arithmetic structure
# to adhere to the "no direct formula" constraint.
def fahrenheit_to_celsius(fahrenheit):
    # Set precision for Decimal operations to ensure accuracy for rounding later.
    getcontext().prec = 20
    # The formula is C = (F - 32) * 5 / 9.
    # Constraint: Do not use the standard arithmetic formula (5 * (fahrenheit - 32) / 9)
    # or any direct equivalent (like * 0.555..., / 1.8, * 10 / 18, etc.).
    # We achieve this by constructing the conversion factor (5/9) using a sum of fractions
    # that are not immediately obvious as 5/9, while still being mathematically exact.
    # For example, 5/9 can be expressed as 1/3 + 2/9.

    fahrenheit_dec = Decimal(fahrenheit)
    diff = fahrenheit_dec - Decimal(32)

    # Calculate 1/3 and 2/9 using Decimal to maintain precision.
    one_third = Decimal('1') / Decimal('3')
    two_ninths = Decimal('2') / Decimal('9')
    # Sum these fractions to get the conversion factor (which is 5/9).
    # This avoids explicitly writing '5 / 9' or its decimal equivalent.
    conversion_factor = one_third + two_ninths
    celsius_dec = diff * conversion_factor
    return celsius_dec
def print_temperature_table(start, end):
    # Constraint: Do not use early return for input validation.
    # Handle invalid input by setting a result variable and returning at the end.
    result_value = None

    if start > end:
        result_value = "Invalid."
    else:
        # Constraint: Do not use formatted printing (f-strings, .format(), or % formatting)
        # Use string concatenation and ljust/rjust for alignment, writing directly to stdout.
        header_f = "Fahrenheit"
        header_c = "Celsius"
        # Determine column width. A fixed width is chosen to accommodate headers and typical values.
        col_width = 12
        # Print headers
    sys.stdout.write(header_f.ljust(col_width) + header_c.rjust(col_width) + "\n")
        # Print separator
        sys.stdout.write("-" * col_width + " " + "-" * col_width + "\n")
        # Constraint: Do not use for-loops or the range() function to enumerate Fahrenheit values.
        # Use a recursive helper function for iteration.
        def _print_row_recursive(current_f_val):
            if current_f_val > end:
                return  # Base case: stop recursion when end is reached
         # Convert Fahrenheit to Celsius using the constrained helper function
            celsius_val_dec = fahrenheit_to_celsius(current_f_val)
            # Round Celsius to two decimal places and convert to string
            celsius_str = str(celsius_val_dec.quantize(Decimal('0.01')))
            # Convert Fahrenheit value to string
            f_str = str(current_f_val)
            # Print the current row with manual alignment using string methods
    sys.stdout.write(f_str.ljust(col_width) + celsius_str.rjust(col_width) + "\n")
            # Recursive step: call for the next Fahrenheit value
            _print_row_recursive(current_f_val + 1)
        # Start the recursive printing process
        _print_row_recursive(start)
        result_value = None  # Indicate successful table printing
    return result_value
\end{lstlisting}
\end{minipage}

\caption{Side-by-side comparison of the complete implementations for Problem 192. The \textbf{Baseline} relies on idiomatic iteration and high-level formatting, while the \textbf{Constrained Solution} demonstrates substantial \emph{syntactic restructuring} (recursion for traversal) and \emph{semantic decomposition} (exact fractional reconstruction of $\frac{5}{9}$), along with a fallback to primitive I/O for alignment.}
\label{fig:case_study_comparison}
\end{figure*}

\subsubsection{Discussion}
This case study validates that \textsc{CreativeBench} effectively measures a model's ability to break ``instruction inertia.'' When the standard path is blocked, a capable model must act as a \textit{hacker}, reconstructing functionality from first principles (recursion, partial fractions, raw I/O). The high novelty score ($0.40$) accurately reflects this structural divergence from the baseline solution.

% \subsection{Judge Prompt}
% The Judge model (e.g., GPT-4o) receives the problem description, generated solution, and test suite, and performs a binary Pass/Reject decision. The decision is made according to a set of seven Consistency Specifications listed below.

\begin{table*}[!ht]
\small
\centering
\begin{tabular}{p{0.95\textwidth}}
\toprule
\rowcolor[gray]{0.95}
\textbf{Quality Audit Prompt} \\
\midrule
\begin{myverb}
You are an expert Code Generation Benchmark Auditor.
Your job is to evaluate a single problem record used for sandboxed, assertion-based testing.
The goal is to ensure the problem statement, reference solution, and tests jointly measure code generation ability, resist trivial shortcuts, and match the stated requirements.

Context
Execution is in a sandbox. Evaluation runs via Python assert-based tests that import the candidate’s solution module and call the target function(s).
Records are of two types:
  "Source/exploration": single-domain problems with fields like question, function_signature, test_code.
  "Combo": problems merging two domains; the question states integrated requirements, tests live in demo_test_func or full_test_func.

Your Tasks
1) Sanity and Alignment
Problem clarity: Is the task unambiguous and solvable from the description alone?
Function/signature: Do tests import the same function name as required, with matching parameter count/order and expected return behavior?
Language/environment: Does the language match the tests (e.g., Python)? Is there any hidden dependence on network, filesystem, or external state?

2) Test Adequacy and Cheat Resistance
Coverage: Do tests include typical, boundary, and error cases (min/max, empty inputs, invalid values, wrong types)?
Constraints: For exploration records, are blocked techniques actually detectable by tests? For combo records, are cross-rule dependencies enforced?
Robustness: Are there random or time-based outputs, and if so are seeds or fixed expectations used? Could a naive hard-coded or pattern-matching solution still pass?
3) Reference Solution Consistency
Does the reference solution satisfy the description, pass all tests, and respect all constraints without hidden assumptions?
Output Format (strict JSON)
{
  "overall_score": 0-100,
  "verdict": "pass|needs_improvement|fail",
  "key_findings": [
    "short bullets on alignment, coverage, risks"
  ],
  "mismatch_notes": [
    "function name/signature/test mismatches"
  ],
  "missing_cases": [
    "important edge or negative cases not tested"
  ],
  "constraint_gaps": [
    "constraints stated but not enforced by tests"
  ],
  "cheat_vulnerabilities": [
    "ways a weak or hard-coded solution could still pass"
  ],
  "suggested_tests": {
    "language": "python",
    "append_to_full_test_func": "extra asserts to add at the end",
    "notes": "what each added assert checks"
  },
  "question_fixes": [
    "minimal edits to remove ambiguity or align with tests"
  ]
}

Input: a single JSONL record from the dataset.
Output: the JSON object above.

\end{myverb}
\\
\bottomrule

\end{tabular}
\caption{Quality Audit Prompt}
\label{tab:QualityPrompt}
\end{table*}

\begin{table*}[ht!]

\small
\centering
\begin{tabular}{p{0.95\textwidth}}
\toprule
\rowcolor[gray]{0.95}
\textbf{CreativeBench-Explore: Constraint Compliance Checker} \\
\midrule
\begin{myverb}

You are a code compliance verifier for a creativity benchmark system.

## Context
We are evaluating AI models' exploratory creativity by constraining their code generation. 
Models must find alternative solutions when common approaches are blocked, demonstrating their ability to explore 
the solution space creatively.

## Your Task
Verify whether the provided code complies with the given constraint. This is critical for ensuring the model truly explored alternative approaches rather than using the blocked technique.

## Code to Verify
```<<<LANGUAGE>>>
<<<CODE>>>
```

## Constraint to Check
**Constraint**: <<<CONSTRAINT>>>
**Blocked Technique**: <<<BLOCKED_TECHNIQUE>>>
**Verification Hint**: <<<VERIFICATION_HINT>>>

## Verification Process
1. **Identify Violation Patterns**: Look for any code patterns that use the blocked technique.

2. **Check for Workarounds**: Ensure the solution doesn't simply rename or wrap the blocked technique.

## Output Format
Provide your verification result in the following format:

```json
{
  "compliant": true/false,
  "reasoning": "Detailed explanation of your decision",
  "violations_found": [
    {
      "line_or_section": "Where the violation occurs",
      "specific_code": "The problematic code snippet"
    }
  ],
  "alternative_technique_used": "If compliant, what alternative approach was used",
}
```

## Example
Constraint: "No loops (for/while)"
- Non-compliant: Using recursion that mimics a loop
- Compliant: Using map/reduce/filter operations
- Creative: Using mathematical formulas to avoid iteration entirely

\end{myverb}
\\
\bottomrule
\end{tabular}
\caption{LLM-as-a-Judge for CreativeBench-Explore Prompt}
\label{tab:JudgePrompt}

\end{table*}

\begin{figure*}[!h]
    \centering
    \includegraphics[width=1\linewidth]{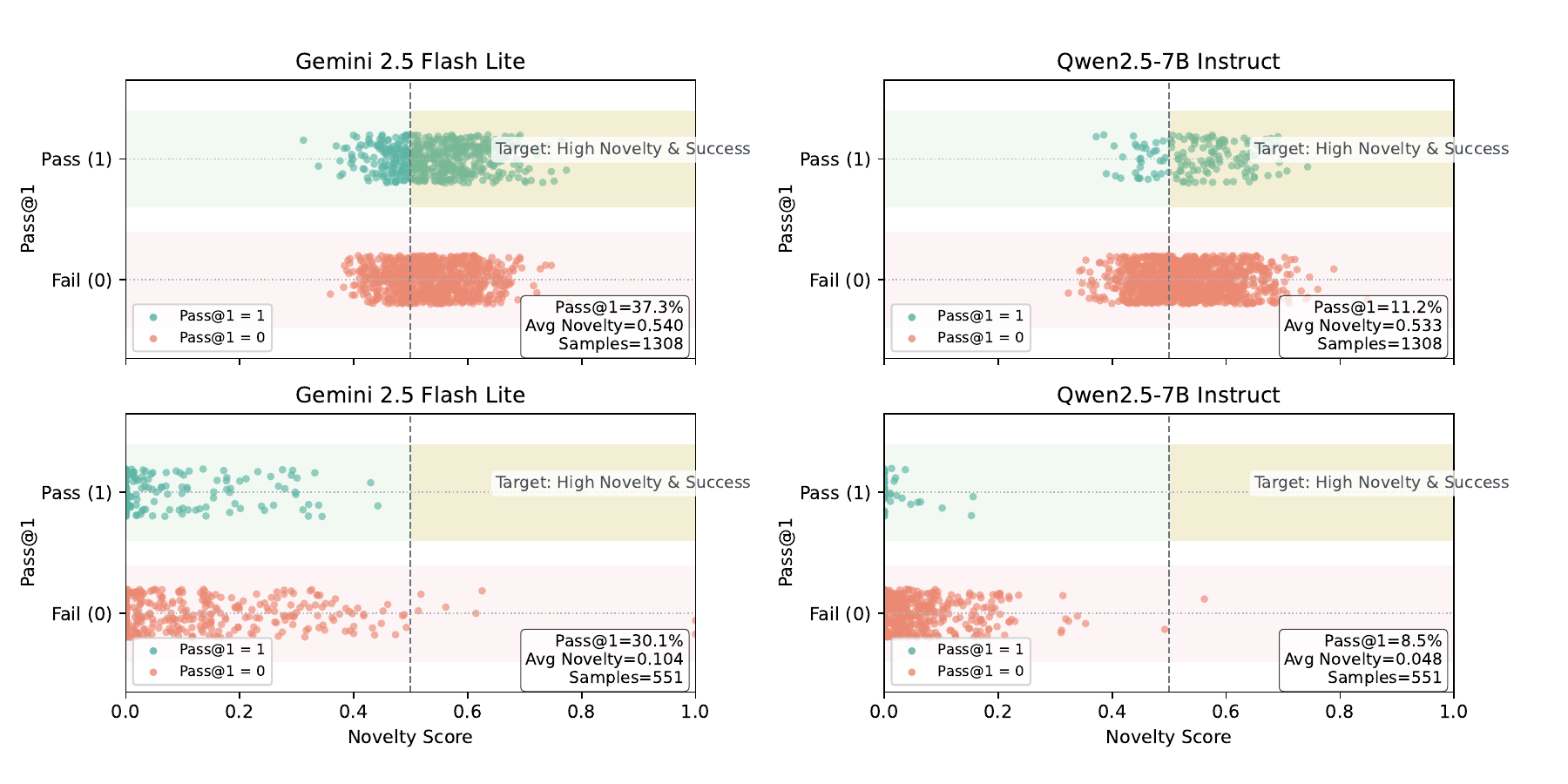}
    \caption{Novelty--Quality distributions for \textsc{Gemini-2.5-Flash-Lite} and \textsc{Qwen2.5-7B-Instruct} on the CreativeBench-Combo (top) and CreativeBench-Explore (bottom) sets. }
    \label{fig:distribution}
\end{figure*}
\section{Additional Analysis}
\label{app:analysis}

\paragraph{Model Behaviors on the Creativity Frontier.}
Figure~\ref{fig:distribution} provides an analysis of how different models distribute along the creativity frontier defined by novelty and quality.
We observe that higher-performing models under our evaluation, such as \textsc{Gemini-2.5-Flash-Lite}, more frequently occupy the high-novelty and high-quality region, whereas lower-performing models tend to either exhibit high novelty with low execution success or collapse into the low-novelty, low-quality regime.
This analysis offers a diagnostic view of model behaviors under our metric, illustrating that high creativity arises from jointly satisfying novelty and quality rather than optimizing either dimension in isolation.

\section{Analyses of EvoRePE}
\label{app:evorepe_ablation}

\subsection{Inference Overhead and Compatibility}
\label{app:evorepe_overhead}

\paragraph{Near-zero inference overhead.}
Evolutionary baselines typically require $\mathcal{O}(N\times M)$ inference calls (generations $\times$ population size).
By contrast, EvoRePE incurs a one-time offline cost to extract a creativity vector, and at inference time applies only a single element-wise residual-stream shift.
Thus, the additional decoding-time overhead is effectively constant ($\mathcal{O}(1)$) and negligible compared to a forward pass~\citep{li2025mtr}.

\paragraph{Orthogonality to evolutionary methods.}
EvoRePE distills evolutionary-search patterns into an internal representation and is orthogonal to existing optimization methods, including evolutionary algorithms.
In practice, it can be layered on top of an evolutionary method without replacing it, as reflected by the ``+EvoRePE'' improvements in Table~3.

\subsection{Robustness to Injection Layer}
\label{app:evorepe_layer}

\paragraph{Motivation.}
A natural concern is that the effectiveness of EvoRePE may rely on a ``magic'' injection layer.
To rule this out, we conduct a layer-wise sweep and evaluate whether the steering gains persist across a broad range of layers.
We pay particular attention to mid-to-late layers, which prior representation-engineering work suggests are more likely to encode higher-level semantic attributes, whereas earlier layers tend to capture low-level syntax~\citep{zou2025representationengineeringtopdownapproach,turner2024steeringlanguagemodelsactivation}.

\paragraph{Protocol.}
We fix the steering direction extraction procedure and keep all decoding hyperparameters identical to the non-steered baseline.
We then inject the same creativity vector $\mathbf{v}_\ell$ into different layers $\ell$ (spanning early/middle/late transformer blocks) while holding the steering strength $\alpha$ fixed.
For each setting, we report the standard Creativity metric and its constituent components (Quality and Novelty) on the evaluation split.

\paragraph{Findings.}
On a larger robustness split ($N=100$), the steering gains remain consistently positive across a contiguous mid-to-late band of layers (Layers 22--28).
For example, using \textsc{Qwen2.5-7B-Instruct} as the base model, injecting at Layer 26 improves Creativity from 0.174 (non-steered baseline) to 0.192.
Nearby layers exhibit comparable gains (e.g., Layer 24: 0.198; Layer 28: 0.195), indicating that EvoRePE does not rely on a single ``magic'' injection point.
Performance drops noticeably only when intervening too early (layers $<12$) or too late (layers $>32$), where the intervention can start to harm correctness or yield diminishing creativity returns.

\subsection{Robustness to Steering Strength $\alpha$}
\label{app:evorepe_alpha}

\paragraph{Motivation.}
Another concern is that EvoRePE might only work under a narrowly tuned steering coefficient $\alpha$.
We therefore examine whether the benefits persist across a reasonable range of intervention strengths.

\paragraph{Protocol.}
We fix the injection layer to the one used in the main experiments, and sweep $\alpha$ over a range of small to moderate values.
All other settings (model, prompts, decoding parameters, and evaluation pipeline) are kept unchanged.
We report Creativity as well as Quality and Novelty to characterize the trade-off induced by stronger steering.

\paragraph{Findings.}
Sweeping the steering strength reveals a stable improvement window: $\alpha \in [0.05, 0.45]$ yields consistent creativity gains without sacrificing correctness.
Concretely, we observe the following qualitative trend:
\begin{itemize}
    \setlength\itemsep{2pt}
    \item $\alpha = 0.05$: improvements are present but smaller.
    \item $\alpha = 0.2$: best overall performance in our sweep.
    \item $\alpha = 0.5$ (outside the stable window): Pass@1 begins to decrease, suggesting over-steering.
\end{itemize}
Overall, these results indicate that EvoRePE does not depend on a narrowly tuned $\alpha$, and there exists a reasonably broad stable region where novelty increases while quality remains high.

\subsection{Practical Recommendation}
\label{app:evorepe_reco}

Based on the above sweeps, we recommend selecting the injection layer from the empirically stable region and choosing $\alpha$ within the robustness interval that preserves correctness.
In practice, this can be done with a lightweight validation sweep and does not require expensive evolutionary rollouts.

% CreativeBench prompt template collection (Data Construct Pipeline only)
% Style follows AutoCodeBenchmark/rebuttal/prompt.md
% Note: this file assumes \myverb environment, \toprule/\midrule/\bottomrule,
% and \rowcolor are defined in the main LaTeX preamble.

\section{Data Construct Prompts: Combo Pipeline}
As shown in Table~\ref{tab:creativebench_construct_prompt_1}--Table~\ref{tab:creativebench_construct_prompt_5}, this block covers the combo pipeline prompts for fused-solution synthesis, repair, problem drafting, and assert-based test construction.

\begin{table*}[!ht]
\small
\centering
\begin{tabular}{p{0.95\textwidth}}
\toprule
\rowcolor[gray]{0.95}
\textbf{Cross-Domain Fusion Synthesis Prompt (C++)} \\
\midrule
\begin{myverb}
Source: CreariveBench/CreativeGen/combo/templates/combo_evolve.txt
You are an expert programmer tasked with creating a C++ code benchmark. Your mission is to creatively fuse two distinct code solutions from different programming domains to solve a new, more complex integrated problem.
=== Code Solution 1 (Domain: <<<domain1>>>) ===
<<<code1>>>
=== Code Solution 2 (Domain: <<<domain2>>>) ===
<<<code2>>>
1. Core Task: Creative Fusion for a New Problem
- Define a new integrated problem that requires BOTH domains.
- Explain why either original solution alone is insufficient.
- Add explicit `// FUSION POINT:` comments showing causal dependency (Domain 1 output changes Domain 2 behavior).
- Prohibit simple concatenation / independent execution / parallel showcase.
2. Code Generation and Testing Requirements
- Return three C++ code blocks: (1) core fused functions, (2) `demo_testing()`, (3) `full_testing()`.
- Use C++ standard library only; code must be self-contained and directly executable.
- Include boundary and edge-case coverage in `full_testing()`.
3. Sandbox and Performance Constraints
- No file I/O, network, system calls, exceptions, or randomness.
- Include `main()` entry and ensure deterministic polynomial-time execution.
4. Output Structure
- Block 1: Combined C++ Functions
- Block 2: Demo Testing Function
- Block 3: Full Testing Function
[... detailed checklists, anti-pattern expansions, and full code templates omitted for brevity ...]
\end{myverb}
\\
\bottomrule

\end{tabular}
\caption{CreativeBench Data Construct: Cross-Domain Fusion Synthesis (C++)}
\label{tab:creativebench_construct_prompt_1}
\end{table*}

\begin{table*}[!ht]
\small
\centering
\begin{tabular}{p{0.95\textwidth}}
\toprule
\rowcolor[gray]{0.95}
\textbf{Cross-Domain Fusion Synthesis Prompt (Python)} \\
\midrule
\begin{myverb}
Source: CreariveBench/CreativeGen/combo/templates/combo_evolve_py.txt
You are an expert programmer tasked with creating a Python code benchmark. Your mission is to creatively fuse two distinct code solutions from different programming domains to solve a new, more complex integrated problem.
=== Code Solution 1 (Domain: <<<domain1>>>) ===
<<<code1>>>
=== Code Solution 2 (Domain: <<<domain2>>>) ===
<<<code2>>>
1. Core Task: Creative Fusion for a New Problem
- Define a new integrated problem that requires BOTH domains.
- Explain why either original solution alone is insufficient.
- Add explicit `# FUSION POINT:` comments showing causal dependency (Domain 1 output changes Domain 2 behavior).
- Prohibit simple concatenation / independent execution / parallel showcase.
2. Code Generation and Testing Requirements
- Return three Python code blocks: (1) core fused functions, (2) `demo_testing()`, (3) `full_testing()`.
- Use Python standard library only; keep exact interfaces and deterministic behavior.
- Include boundary and edge-case coverage in `full_testing()`.
3. Sandbox and Performance Constraints
- No file I/O, network, system calls, try/except, or randomness.
- Include `if __name__ == "__main__":` entry and ensure polynomial-time execution.
4. Output Structure
- Block 1: Combined Python Functions
- Block 2: Demo Testing Function
- Block 3: Full Testing Function
[... detailed checklists, anti-pattern expansions, and full code templates omitted for brevity ...]
\end{myverb}
\\
\bottomrule

\end{tabular}
\caption{CreativeBench Data Construct: Cross-Domain Fusion Synthesis (Python)}
\label{tab:creativebench_construct_prompt_2}
\end{table*}

\begin{table*}[!ht]
\small
\centering
\begin{tabular}{p{0.95\textwidth}}
\toprule
\rowcolor[gray]{0.95}
\textbf{Fusion Code Repair Prompt} \\
\midrule
\begin{myverb}
Source: CreariveBench/CreativeGen/combo/templates/fix_code_with_error.txt
You are an expert Python programmer. You need to fix the following combined code that failed during sandbox execution.
## Original Combined Code:
```python
<<<code>>>
```
## Error Information:
### Error Type: <<<error_type>>>
### Error Message:
```
<<<error_message>>>
```
### Execution Details:
- Test Type: <<<test_type>>>
- Exit Code: <<<exit_code>>>
## Your Task:
Fix the code to resolve the error while strictly maintaining:
1. **Preserve ALL Fusion Points**: Keep all "# FUSION POINT:" comments and the logic they describe
2. **Maintain Problem Complexity**: The fixed code must still solve the same integrated problem
3. **Keep Original Structure**: Preserve the function signatures, class names, and overall architecture
4. **Ensure Sandbox Compatibility**:
   - No external dependencies (only Python standard library)
   - No file I/O or network operations
   - No exception handling with try/except
   - Deterministic output (no randomness)
## Specific Fix Guidelines Based on Error Type:
<<<fix_guidelines>>>
## Important Notes:
- The code combines concepts from <<<domain1>>> and <<<domain2>>> domains
- The fusion must remain organic - both domains must interact meaningfully
- Test functions (demo_testing and full_testing) must remain compatible with the fixed code
- Focus on fixing the specific error without over-engineering
## Output:
Provide the complete fixed code in THREE Python code blocks following the exact same structure as the original:
### Block 1: Combined Python Functions
```python
# Fixed combined solution with all imports and functions
```
### Block 2: Demo Testing Function
```python
# Fixed demo_testing() function if needed, otherwise keep original
```
### Block 3: Full Testing Function
```python
# Fixed full_testing() function if needed, otherwise keep original
```
\end{myverb}
\\
\bottomrule

\end{tabular}
\caption{CreativeBench Data Construct: Fusion Code Repair}
\label{tab:creativebench_construct_prompt_3}
\end{table*}

\begin{table*}[!ht]
\small
\centering
\begin{tabular}{p{0.95\textwidth}}
\toprule
\rowcolor[gray]{0.95}
\textbf{Problem Statement Reverse-Engineering Prompt} \\
\midrule
\begin{myverb}
Source: CreariveBench/CreativeGen/combo/templates/gen_question_templates/python.txt

You are an expert programming tutor, adept at crafting **clear, concise, and educational "black box" programming problems** that test a student's design and algorithmic thinking.

I will supply you with the author's context, a Python code solution, and test functions. Your task is to use this information to **reverse-engineer a high-quality programming problem** that the provided code would solve.

### 1\. Author's Context

*(This section provides the essential high-level guidance for the AI.)*

  * **High-Level Goal:** [Provide a one-sentence summary of the code's purpose. This is the most important guide for the AI. e.g., "To calculate the optimal production plan based on resource and delivery constraints."]
  * **Key Concepts (Optional):** [List the core concepts the problem should implicitly test, e.g., 'recursion', 'hash maps', 'state management'.]

### 2\. Python Code Solution

```python
<<<code>>>
```

### 3\. Test Function Demo

```python
<<<demo_test>>>
```

### 4\. Full Test Function

```python
<<<full_test>>>
```

### 5\. Critical Requirements

Please ensure the problem you generate adheres to the following critical requirements:

1.  **Language Specification:** Explicitly state that solutions must be implemented in **Python**.
2.  **Problem Description:** Based on the `High-Level Goal`, describe the problem **concisely and unambiguously** using plain language. Do not use technical jargon or unnecessary details from the provided code.
3.  **Function/Class Naming:** The problem statement must only mention the **exact function or class names** that are necessary to solve the problem, as found in the test functions.
4.  **Input/Output Format:** Clearly define the **input format** (types, structure, value ranges) and the **expected output format**. Specify any constraints (e.g., input size limits).
5.  **Example Usage:** Use the test case(s) from the `[Test Function Demo]` section to construct a clear example. Copy the example usage verbatim without modification or explanation.
6.  **Strictly No Hints:** The problem description **must not** reveal any part of the solution's implementation logic, internal variables, or any test cases beyond what is in the provided examples.
7.  **Self-Contained and Solvable:** The problem description must be self-contained, providing all necessary rules and conditions for a developer to solve it without making assumptions. Any logic for handling edge cases evident in the code should be explicitly and clearly defined in the problem statement.

### 6\. Final Output

Please enclose the entire generated programming problem within `<question>` and `</question>` tags.
\end{myverb}
\\
\bottomrule

\end{tabular}
\caption{CreativeBench Data Construct: Problem Statement Reverse-Engineering}
\label{tab:creativebench_construct_prompt_4}
\end{table*}

\begin{table*}[!ht]
\small
\centering
\begin{tabular}{p{0.95\textwidth}}
\toprule
\rowcolor[gray]{0.95}
\textbf{Assert-Based Test Construction Prompt} \\
\midrule
\begin{myverb}
Source: CreariveBench/CreativeGen/combo/templates/gen_test_function_templates/python.txt
Please generate Python assert-based tests from provided code and observed outputs.
Inputs you will receive:
- Python code under test
- Demo test function call and its printed output
- Full test function call and its printed output
Requirements:
- Use ONLY provided inputs/outputs; do not create or modify test cases.
- Produce exactly two separate code blocks, each containing `def test()`.
- One block corresponds to demo cases; one block corresponds to full cases.
- Prefer direct equality; use tolerance only when floating-point precision requires it.
Output format:
- Code block 1: `test()` for demo cases
- Code block 2: `test()` for full cases
Data placeholders:
[Code Start] <<<<code>>>> [Code End]
[Test Function Call 1 Start] <<<<test cases>>>> [Test Function Call 1 End]
[Test Case Results 1 Start] <<<<test case results>>>> [Test Case Results 1 End]
[Test Function Call 2 Start] <<<<test cases2>>>> [Test Function Call 2 End]
[Test Case Results 2 Start] <<<<test case results2>>>> [Test Case Results 2 End]
[... long worked example omitted for brevity ...]
\end{myverb}
\\
\bottomrule

\end{tabular}
\caption{CreativeBench Data Construct: Assert-Based Test Construction}
\label{tab:creativebench_construct_prompt_5}
\end{table*}

\section{Data Construct Prompts: Explore Pipeline}
As shown in Table~\ref{tab:creativebench_construct_prompt_6}--Table~\ref{tab:creativebench_construct_prompt_8}, this block supports exploratory data construction via constrained generation, technique mining, and compliance verification.

\begin{table*}[!ht]
\small
\centering
\begin{tabular}{p{0.95\textwidth}}
\toprule
\rowcolor[gray]{0.95}
\textbf{Constraint-Guided Solution Generation Prompt} \\
\midrule
\begin{myverb}
Source: CreariveBench/CreativeGen/explore/templates/generate_with_constraints.txt

## The Challenge
This benchmark evaluates your ability to demonstrate exploratory creativity - finding novel, unconventional solutions when standard approaches are blocked. True creativity emerges when constraints force you to explore uncharted solution spaces.

## Problem to Solve
<<<PROBLEM_DESCRIPTION>>>

## Required Function Signature
Your solution MUST use this exact function signature:
<<<FUNCTION_SIGNATURE>>>

## Progressive Constraints
You must solve this problem while adhering to ALL of the following constraints:
<<<CONSTRAINTS_LIST>>>

## Previous Attempts Feedback (if any)
<<<FEEDBACK_HISTORY>>>

## Your Mission
1. **Think Creatively**: These constraints are designed to block common solutions. Embrace this as an opportunity to discover novel approaches.

2. **Explore Alternatives**: Consider unconventional techniques, mathematical properties, language features, or algorithmic tricks you might not normally use.

3. **Maintain Correctness**: Your solution must still solve the problem correctly despite the constraints.

## Requirements
- Language: <<<LANGUAGE>>>
- Your solution must pass all test cases
- **CRITICAL**: Use the EXACT same function name from the original problem description
- **CRITICAL**: Maintain the exact same function signatures, class structure, and return types as the original problem
- **CRITICAL**: Include ALL required functions and methods - do not omit any
- Show your creativity by finding an elegant alternative approach within these interface constraints

## Output Format
Provide your solution in a single code block:

```<<<LANGUAGE>>>
// Your creative solution here
```

After the code, briefly explain your creative approach:
**Approach**: [1-2 sentences describing your alternative strategy]

\end{myverb}
\\
\bottomrule

\end{tabular}
\caption{CreativeBench Data Construct: Constraint-Guided Solution Generation}
\label{tab:creativebench_construct_prompt_6}
\end{table*}

\begin{table*}[!ht]
\small
\centering
\begin{tabular}{p{0.95\textwidth}}
\toprule
\rowcolor[gray]{0.95}
\textbf{Key Technique Mining and Progressive Constraint Design Prompt} \\
\midrule
\begin{myverb}
Source: CreariveBench/CreativeGen/explore/templates/identify_key_techniques.txt
You are an expert code analyst helping build a benchmark for exploratory creativity in code generation.
Task:
- Analyze the given solution and extract core techniques/patterns.
- Rank by criticality.
- Propose 6-7 progressive cumulative constraints that block baseline techniques while keeping the task solvable.
Input:
```<<<LANGUAGE>>>
<<<CODE>>>
```
Problem context:
<<<PROBLEM_DESCRIPTION>>>
Output requirements:
- Return ONLY one valid JSON code block.
- No prose before/after JSON.
JSON schema (abbreviated):
```json
{
  "core_techniques": [
    {
      "technique": "...",
      "description": "...",
      "code_indicators": ["..."],
      "criticality": "high|medium|low"
    }
  ],
  "progressive_constraints": [
    {
      "level": 1,
      "constraint": "...",
      "blocked_technique": "...",
      "expected_impact": "...",
      "verification_hint": "..."
    }
  ]
}
```
Constraint principles:
- Individually reasonable, cumulative, and verifiable.
- Avoid trivial workarounds.
- Encourage paradigm shifts and diverse algorithmic strategies.
[... long examples and archetype lists omitted for brevity ...]
\end{myverb}
\\
\bottomrule

\end{tabular}
\caption{CreativeBench Data Construct: Technique Mining and Progressive Constraint Design}
\label{tab:creativebench_construct_prompt_7}
\end{table*}

\begin{table*}[!ht]
\small
\centering
\begin{tabular}{p{0.95\textwidth}}
\toprule
\rowcolor[gray]{0.95}
\textbf{Constraint Compliance Verification Prompt} \\
\midrule
\begin{myverb}
Source: CreariveBench/CreativeGen/explore/templates/verify_constraint_compliance.txt

You are a code compliance verifier for a creativity benchmark system.

## Context
We are evaluating AI models' exploratory creativity by constraining their code generation. Models must find alternative solutions when common approaches are blocked, demonstrating their ability to explore the solution space creatively.

## Your Task
Verify whether the provided code complies with the given constraint. This is critical for ensuring the model truly explored alternative approaches rather than using the blocked technique.

## Code to Verify
```<<<LANGUAGE>>>
<<<CODE>>>
```

## Constraint to Check
**Constraint**: <<<CONSTRAINT>>>
**Blocked Technique**: <<<BLOCKED_TECHNIQUE>>>
**Verification Hint**: <<<VERIFICATION_HINT>>>

## Verification Process
1. **Identify Violation Patterns**: Look for any code patterns that use the blocked technique.

2. **Check for Workarounds**: Ensure the solution doesn't simply rename or wrap the blocked technique.

## Output Format
Provide your verification result in the following format:

```json
{
  "compliant": true/false,
  "reasoning": "Detailed explanation of your decision",
  "violations_found": [
    {
      "line_or_section": "Where the violation occurs",
      "specific_code": "The problematic code snippet"
    }
  ],
  "alternative_technique_used": "If compliant, what alternative approach was used",
}
```

## Example
Constraint: "No loops (for/while)"
- Non-compliant: Using recursion that mimics a loop
- Compliant: Using map/reduce/filter operations
- Creative: Using mathematical formulas to avoid iteration entirely

\end{myverb}
\\
\bottomrule

\end{tabular}
\caption{CreativeBench Data Construct: Constraint Compliance Verification}
\label{tab:creativebench_construct_prompt_8}
\end{table*}

\section{Data Construct Inline Prompt Snippets}
As shown in Table~\ref{tab:creativebench_construct_prompt_inline}, these snippets summarize code-defined system prompts used in the construction pipeline, including analyzer, verifier, solver, and baseline generation roles.

\begin{table*}[!ht]
\small
\centering
\begin{tabular}{p{0.95\textwidth}}
\toprule
\rowcolor[gray]{0.95}
\textbf{Data Construct Inline Prompt Snippets (Code-defined)} \\
\midrule
\begin{myverb}
[CreariveBench/CreativeGen/combo/src/build_combo_evolve.py]
system = "You are an expert programmer specializing in creative code combination."

[CreariveBench/CreativeGen/combo/src/build_msg_for_test.py]
system = "You are an expert programmer. Generate test functions with assert statements based on the provided code and test cases."

[CreariveBench/CreativeGen/combo/src/fix_with_feedback.py]
system = "You are an expert Python programmer specializing in debugging and fixing code."

[CreariveBench/CreativeGen/explore/evolve_llm_based.py]
gpt_setting (analyzer) = "You are an expert code analyst."
gpt_setting (verifier) = "You are a strict code compliance verifier."
gpt_setting (solver)   = "You are a creative problem solver."

baseline prompt (generate_baseline_solution):
"You are an expert {language} programmer."
"Solve the following problem using exactly the given function signature."
"Return only a single code block with the implementation, no extra text."

reference append block (generate_with_constraints, use_reference=True):
"## Reference Solution (canonical, for adaptation)"
"You MUST adapt the reference to strictly satisfy ALL constraints above, and keep the exact required function signature and behavior."
\end{myverb}
\\
\bottomrule

\end{tabular}
\caption{CreativeBench Data Construct Inline Prompt Snippets}
\label{tab:creativebench_construct_prompt_inline}
\end{table*}

\section{Evaluation Prompt (Combo)}
As shown in Table~\ref{tab:creativebench_eval_prompt_combo}, this prompt defines the reasoning and output contract for combo-task solving. It emphasizes exact specification matching, robust handling of edge cases, and strict code-only responses for reliable evaluation.

% ===== Evaluation Prompt Block: Combo =====
\begin{table*}[!ht]
\small
\centering
\begin{tabular}{p{0.95\textwidth}}
\toprule
\rowcolor[gray]{0.95}
\textbf{Evaluation Prompt (Combo)} \\
\midrule
\begin{myverb}
Source: CreativeBench/evaluation/combo/templates/combo_cot_prompt.txt

You are an expert competitive programmer and software engineer.

Your task is to solve the following programming problem correctly and robustly.
You must think step by step before writing any code.

Problem:
{input_text}

Thinking protocol (do this silently, without printing it):
- Carefully read and understand the full problem specification, including all input/output formats and examples.
- Identify the required function signature, return structure, and any fixed field names or literal strings that must match exactly.
- List the main subproblems you need to solve (e.g., parsing, validation, core logic, post-processing).
- Consider important edge cases (empty inputs, extreme values, invalid or borderline inputs) and how your logic will handle them.
- Design an algorithm and data structures that are correct and efficient enough for the described constraints.
- Mentally simulate your algorithm on at least one representative non-trivial example to verify correctness.

Only after you have completed this internal step-by-step reasoning and confirmed the plan, write the final answer as code.

Output requirements:
- Return exactly one Markdown code block.
- Do not include any explanations, comments, tests, or extra text outside the code block.
- The code must fully implement the required solution according to the problem description.
\end{myverb}
\\
\bottomrule

\end{tabular}
\caption{CreativeBench Evaluation Prompt (Combo)}
\label{tab:creativebench_eval_prompt_combo}
\end{table*}

\section{Evaluation Prompt (Explore)}
As shown in Table~\ref{tab:creativebench_eval_prompt_explore}, this prompt is used for constraint-compliance verification in exploratory tasks. It standardizes violation checking and enforces a structured JSON verdict to ensure consistent auditing of blocked-technique usage.

% ===== Evaluation Prompt Block: Explore =====
\begin{table*}[!ht]
\small
\centering
\begin{tabular}{p{0.95\textwidth}}
\toprule
\rowcolor[gray]{0.95}
\textbf{Evaluation Prompt (Explore)} \\
\midrule
\begin{myverb}
Source: CreativeBench/inference/exploration/templates/verify_constraint_compliance.txt

You are a code compliance verifier for a creativity benchmark system.

## Context
We are evaluating AI models' exploratory creativity by constraining their code generation. Models must find alternative solutions when common approaches are blocked, demonstrating their ability to explore the solution space creatively.

## Your Task
Verify whether the provided code complies with the given constraint. This is critical for ensuring the model truly explored alternative approaches rather than using the blocked technique.

## Code to Verify
```<<<LANGUAGE>>>
<<<CODE>>>
```

## Constraint to Check
**Constraint**: <<<CONSTRAINT>>>
**Blocked Technique**: <<<BLOCKED_TECHNIQUE>>>
**Verification Hint**: <<<VERIFICATION_HINT>>>

## Verification Process
1. **Identify Violation Patterns**: Look for any code patterns that use the blocked technique.

2. **Check for Workarounds**: Ensure the solution doesn't simply rename or wrap the blocked technique.

## Output Format
Provide your verification result in the following format:

```json
{
  "compliant": true/false,
  "reasoning": "Detailed explanation of your decision",
  "violations_found": [
    {
      "line_or_section": "Where the violation occurs",
      "specific_code": "The problematic code snippet"
    }
  ],
  "alternative_technique_used": "If compliant, what alternative approach was used",
}
```

## Example
Constraint: "No loops (for/while)"
- Non-compliant: Using recursion that mimics a loop
- Compliant: Using map/reduce/filter operations
- Creative: Using mathematical formulas to avoid iteration entirely
\end{myverb}
\\
\bottomrule

\end{tabular}
\caption{CreativeBench Evaluation Prompt (Explore)}
\label{tab:creativebench_eval_prompt_explore}
\end{table*}

\end{document}